\documentclass[12pt,reqno]{article}
\newif\ifclean
\cleantrue  
\usepackage{graphicx}
\usepackage{cite}
\usepackage{amsmath}
\usepackage{amssymb}
\usepackage{subcaption}
\usepackage{hyperref}
\usepackage[letterpaper,margin=0.75in]{geometry}
\graphicspath{{{./}}}

\ifclean
\newcommand{\COMMENT}[1]{{}}
\newcommand{\QUESTION}[1]{{}}
\newcommand{\TODO}[1]{{}}
\else
\usepackage[usenames,dvipsnames]{xcolor}
\newcommand{\COMMENT}[1]{\textcolor{cyan}{{[ \sc{#1} ]}}\\} 
\newcommand{\QUESTION}[1]{\textcolor{cyan}{{[QUESTION: \it{#1} ]}}\\} 
\newcommand{\TODO}[1]{\textcolor{red}{TODO: {\bf{#1}}}\\} 

\newcommand{\red}[1]{\textcolor{red}{{#1}}}
\newcommand{\caution}{\red{\bf Draft: \today. Do not distribute.}}
\fi

\newcommand{\aref}[1]{App.\,\ref{#1}}
\newcommand{\fref}[1]{Fig.\,\ref{#1}}

\newcommand{\eref}[1]{Eq.\,(\ref{#1})}

\newcommand{\sref}[1]{Sec.\!~\ref{#1}}

\newcommand{\cref}[1]{Ref.\,\cite{#1}}

\newcommand{\ie}{{\it i.e.}\!\, }
\newcommand{\eg}{{\it e.g.}\!\, }

\newcommand{\etal}{{\it et al.} }

\newcommand{\Cbb}{\mathbb{C}}
\newcommand{\Vbb}{\mathbb{V}}

\newcommand{\eb}{\mathbf{e}}

\newcommand{\vb}{\mathbf{v}}
\newcommand{\xb}{\mathbf{x}}

\newcommand{\nb}{\mathbf{n}}
\newcommand{\qb}{\mathbf{q}}
\renewcommand{\sb}{\mathbf{s}}

\newcommand{\Db}{\mathbf{D}}
\newcommand{\Eb}{\mathbf{E}}
\newcommand{\Fb}{\mathbf{F}}

\newcommand{\Lb}{\mathbf{L}}

\newcommand{\Ib}{\mathbf{I}}
\newcommand{\Rb}{\mathbf{R}}
\newcommand{\Sb}{\mathbf{S}}
\newcommand{\Tb}{\mathbf{T}}
\newcommand{\Xb}{\mathbf{X}}

\newcommand{\As}{\mathsf{A}}

\newcommand{\Xs}{\mathsf{X}}
\newcommand{\Ss}{\mathsf{S}}
\newcommand{\Zs}{\mathsf{Z}}

\newcommand{\Ws}{\mathsf{W}}

\newcommand{\Vs}{\mathsf{V}}
\newcommand{\bs}{\mathsf{b}}

\newcommand{\Fs}{\mathsf{F}}

\newcommand{\epsilonb}{{\boldsymbol{\epsilon}}}
\newcommand{\kappab}{{\boldsymbol{\kappa}}}

\newcommand{\sigmab}{{\boldsymbol{\sigma}}}
\newcommand{\phib}{{\boldsymbol{\phi}}}

\newcommand{\abs}{{\operatorname{abs}}}
\newcommand{\tr}{{\operatorname{tr}}}

\newcommand{\dev}{{\operatorname{dev}}}

\newcommand{\grad}{\boldsymbol{\nabla}}
\newcommand{\dV}{\mathrm{d}V}

\newcommand{\vol}{\mathrm{v}}
\newcommand{\partialb}{\boldsymbol{\partial}}

\newcommand{\Conv}{\operatorname{Conv}}
\newcommand{\Pool}{\operatorname{Pool}}
\newcommand{\diffpool}{{\it DiffPool}}
\newcommand{\mincutpool}{{\it MinCutPool}}
\newcommand{\softmax}{{$\operatorname{softmax}$}}

\title{\bf Deep learning and multi-level featurization of graph representations of microstructural data}
\author{
Reese Jones,\footnote{Corresponding author: \tt rjones@sandia.gov} \,
Cosmin Safta,
Ari Frankel
\\
{\it Sandia National Laboratories, Livermore, CA 94551}
}

\begin{document}
\ifclean
\date{}
\else
\date{\caution}
\fi

\maketitle{}

\begin{abstract}
Many material response functions depend strongly on microstructure, such as inhomogeneities in phase or orientation.
Homogenization presents the task of predicting the mean response of a sample of the microstructure to external loading for use in subgrid models and structure-property explorations.
Although many microstructural fields have obvious segmentations, learning directly from the graph induced by the segmentation can be difficult because this representation does not encode all the information of the full field.
We develop a means of deep learning of hidden features on the reduced graph given the native discretization and a segmentation of the initial input field.
The features are associated with regions represented as nodes on the reduced graph.
This reduced representation is then the basis for the subsequent multi-level/scale graph convolutional network model.
There are a number of advantages of reducing the graph before fully processing with convolutional layers it, such as interpretable features and efficiency on large meshes.
We demonstrate the performance of the proposed network relative to convolutional neural networks operating directly on the native discretization of the data using three physical exemplars.
\end{abstract}

\section{Introduction}\label{sec:intro}

Newly developed graph neural networks (GNNs)~\cite{hamilton2017representation,bronstein2017geometric,battaglia2018relational}, in particular convolutional graph neural networks, have been shown to be effective in a variety of classification and regression tasks.
Recently they have been applied to physical problems~\cite{xie2018crystal,vlassis2020geometric} where they can accommodate unstructured and hierarchical data naturally.
Analogous to pixel-based convolutional neural networks (CNNs), ``message passing''~\cite{gilmer2017neural} graph convolutional neural networks (GCNNs)~\cite{defferrard2016convolutional,kipf2016semi} employ convolutional operations to achieve a compact parameter space by exploiting correlations in the data through connectivity defined by adjacency on the source discretization.
Frankel \etal~\cite{frankel2022mesh} and others~\cite{dominguez2017towards,ogoke2020graph,pfaff2020learning,shi2022gnn} derive the information transmission graph directly from the connectivity of the discretization, computational grid or mesh based on the assumption the physical interactions are local.
Some obvious advantages of applying convolutions to the discretization graph are that: general mesh data can be handled without interpolation to a structured grid, the discretization can be conformal to the microstructure, periodic boundary conditions can be handled without padding, and topological irregularities can be accommodated without approximations.
In this approach the kernels and number of parameters are similar for a pre-selected reduction of the representation, \eg based on the grains in a polycrystal \cite{vlassis2020geometric}, but the size of the adjacency can be prohibitive.
Naively constructed and applied, these mesh-based graph models can operate very large graphs; however, graph reduction is important for efficiency and can promote learning \cite{bianchi2020spectral}.
While physical problems with short- (e.g interface) and long-range (e.g. elastic) interactions are ubiquitous in engineering and materials science, treating the longer scales with convolutions on the discretization graph can be inefficient or ineffective.

Treating aggregated data via graph pooling based on data clustering is a long-standing field of research with broad applications, \eg image classification~\cite{yu2003object}, mesh decomposition~\cite{karypis1998fast}, and chemical structure~\cite{ryu2018deeply}.
Since clustered data rarely has a regular topology, graph based networks are natural representations.
In particular, compared to a CNN operating on a pixelized image, a GCNN operating on an unstructured mesh has less well-defined spatial locality and inherits the varying neighborhood size of the source data.
Akin to topologically localized convolutions supplanting spectral-based convolutions~\cite{defferrard2016convolutional,kipf2016variational} and spectral pre-processing~\cite{do1998shape} in the literature, spectral clustering based on the primary eigenvectors of the graph Laplacian has been largely superseded by less expensive, more easily differentiable techniques, some of which connect to spectral clustering.
Dhillon \etal~\cite{dhillon2007weighted} showed the equivalence of spectral clustering and kernel $k$-means clustering which allowed expensive eigenvalue problems to be reframed in terms of trace maximization objectives.
These objectives include maximizing in-cluster links/edges and minimizing cuts, \ie number of links, between any  cluster and remainder of the graph.
Typically these objectives are normalized relative to cluster size (number of nodes) or degree (sum of non-self connections/adjacency) but are ambivalent to the data on the graph.
More recently, graph-based neural nets, such as \diffpool\ \cite{ying2018hierarchical} and \mincutpool\ \cite{bianchi2020spectral}, have been designed to take into account the data on the graph in soft clustering, trainable pooling operations.
Ying \etal~\cite{ying2018hierarchical} developed \diffpool\ to enable a hierarchical treatment of graph structure.
\diffpool\ uses a GNN for the ultimate classification task and another with a \softmax\ output for the intermediary pooling task.
The separate GNNs learn a soft, in the sense not binary and disjoint, assignment of nodes in the input graph to those in a smaller graph as well as derivative features on the smaller embedded graph.
Due to non-convexity of the clustering objective, an auxiliary entropy loss is used to regularize the training.
Bianchi, Grattarola and Alippi \cite{bianchi2020spectral} developed \mincutpool\ based on a degree normalized objective of minimizing edges between any cluster and the remainder of the graph.
They relaxed the objective of finding a binary, disjoint cluster assignment to reduce the computational complexity by recasting the problem to a continuous analog.
Ultimately they proposed a network similar to \diffpool\ albeit with the \softmax\ assignment GNN replaced by a \softmax\ multilayer perceptron (MLP) and different loss augmentation designed to promote orthogonality of the clustering matrix mapping graph nodes to clusters.
Grattarola \etal~\cite{grattarola2021understanding} also generalized the myriad approaches to pooling and clustering on graphs with their select-reduced-connect abstraction of these operations.

Graph convolutional neural networks can be particularly opaque in how they achieve accurate models.
To interpret what convolution networks are learning through their representations, filter visualization and activation~\cite{samek2016evaluating,qin2018convolutional,yosinski2015understanding}, saliency maps~\cite{simonyan2013deep}, sensitivity, attention based and related techniques~\cite{bodria2021benchmarking,zhang2021survey} have been developed.
These techniques have been largely applied to the ubiquitous commercial classification tasks, for instance Yosinski \etal~\cite{yosinski2015understanding} demonstrated how activations at the deepest layers of a CNN correspond to obvious features in the original image used in classification, \eg faces.
Some of these techniques have enabled a degree of filter interpretation by translating the filter output so that the learned features are obvious by inspection.
For pixel-based CNNs deconvolution and inversion of the input-output map~\cite{qin2018convolutional}, as well as guided/regularized optimization~\cite{yosinski2015understanding} have produces some insights.
Some other methods rely on clustering, for example, Local Interpretable Model-agnostic Explanations (LIME) \cite{ribeiro2016model} is based on segmentation and perturbation of the cluster values.
In general, clustering simplifies the input-output map created by the convolution filter and therefore the complexity of the image-to-image convolutional transformation.

Given that in many physical homogenization problems the segmentation of the domain based on specific features, e.g. phase or orientation, of the constituents is readily accessible but the informative features on the sub-domains are not, we propose a means of constructing a graph-based representation where the features on the nodes representing the sub-domains are learned/not pre-selected while the clustered topology is known.
The concept resembles the reduction-prolongation operations of algebraic multi-grid approaches \cite{ruge1987algebraic}, the architecture of graph U-nets \cite{gao2019graph} and the combination of multigrid with pixel-based CNNs \cite{he2019mgnet}.
Beyond developing a GCNN architecture to efficiently reduce data on the native discretization graph to logical clustering without manual feature engineering, a secondary goal is to provide some insight into how the resulting models learn an accurate representation of the data.

This paper is organized as follows.
In \sref{sec:problem} we describe the homogenization problems and in \sref{sec:data_sets} we present the two physical exemplars we model with GCNNs.
\sref{sec:architecture} describes the proposed architecture and contrast it with traditional CNN and GCNN-based architectures.
\sref{sec:interpret} outlines the techniques we use to interpret the flow of information in the networks.
Then in \sref{sec:results} we demonstrate the performance and interpretation of the proposed framework.
With the idea that that the simplest accurate model is likely to be the most interpretable, we compare the proposed deep reduced GCNN to analogous architectures based on previous GCNN \cite{frankel2022mesh} and CNN \cite{frankel2019oligocrystals} representations.
We conclude in \sref{sec:conclusion} with a discussion of the proposed architecture in the context to soft, learned clustering, and extensions left for future work.

\section{Homogenization} \label{sec:problem}

Homogenization of the physical response of a sample with complex structure represents a class of problems with broad physical applications, such as creating sub-grid models of materials with multiple phases, orientations or other microstructural aspects.
For each particular physical problem the input is a sample of the detailed structure in a regular domain and the output is the average or total response of the sample subject to external loading.
Samples that are of a size that is representative of the large sample limit are called representative volume elements (RVEs) and those that are smaller, where statistical variations are apparent in the output, are called statistical volume elements (SVEs) \cite{mura2013micromechanics,nemat2013micromechanics}; this distinction is clearly a matter of degree.

\begin{figure}[htb!]
\centering
\includegraphics[width=0.45\textwidth]{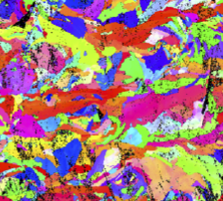}
\caption{Electron backscatter diffraction (EBSD) image of a polycrystalline metal (courtesy of Brad Boyce, Sandia).
Colors indicate the crystal orientations.
}
\label{fig:ebsd}
\end{figure}

For example, the volume average of stress evolution $\sigmab(t)$ as a function of (boundary loading) strain $\bar{\epsilonb}(t)$ is often of interest in developing material models.
The boundary value problem over the sample gives a field, in this example stress $\sigmab(\Xb,t)$, as a function of position $\Xb$ and time $t$:
\begin{equation}
\sigmab(\Xb,t) = \sigmab(\phib(\Xb),\bar{\epsilonb}(t)) ,
\end{equation}
where $\phib(\Xb)$ is a field describing the initial microstructure, and $\epsilonb(t)$ is an external loading applied through the boundary conditions.
Homogenization aims to represent the average response over the sample domain $\Omega$:
\begin{equation}
\bar{\sigmab}(t) = \frac{1}{V} \int_\Omega \sigmab(\Xb,t) \, \dV
\end{equation}
as a function of $\phib(\Xb)$ and $\bar{\epsilonb}(t)$.
Here $V$ is the volume of the sample domain $\Omega$.
The field $\phib(\Xb)$ can be interpreted as a (potentially multichannel) image resulting from, for instance, computed tomography and/or electron backscatter diffraction (refer to \fref{fig:ebsd}).
For this work we assume that $\phib(\Xb)$ over the sample domain $\Omega$ can be readily segmented into disjoint regions $\Omega_K$ that each have a uniform value $\phib_K$, as in \fref{fig:ebsd}.
The discretization necessary to resolve the fields leads to the number of discretization cells $N_\text{cells}$ being much greater than the number of regions $N_\text{clusters}$ in general.

Homogenization traditionally has relied on analytical models based on simplifying assumptions.
Mixture rules where individual contributions of the constituents of the microstructure are summed and weighted by their volume fractions, $\chi_K = V_K / V$, have generally been utilized in approximate analytical homogenization.
For instance, the approximation
\begin{equation} \label{eq:ave_stress}
\bar{\sigmab}
= \frac{1}{V} \sum_K \int_{\Omega_K} \sigmab \, \dV
\approx \sum_K \chi_K \sigmab_K
\end{equation}
where $\sigmab_K = \sigmab(\phib_K,\epsilonb_K)$ is widely employed.
Note that mixture rules such as these typically pair extensive partitioning variables, such as $\chi_K$, with intensive variables, such as $\phib_K$.
The approximation \eref{eq:ave_stress} leads to an estimate of the elastic moduli $\Cbb \equiv \partialb_\epsilonb \sigmab$:
\begin{equation}
\bar{\Cbb} \approx \sum_K \chi_K \Cbb_K ,
\end{equation}
assuming the strain $\epsilonb$ is homogeneous in the sample.
Homogeneous flux or homogeneous field gradient are typical simplifying assumptions for homogenizing gradient-driven fluxes such as stress and heat flux.
These complementary limiting cases can be combined, as in the classic Hill average estimator \cite{hill1952elastic} for the elastic modulus tensor:
\begin{equation} \label{eq:Hill}
\bar{\Cbb} =
\frac{1}{2} \left(\sum_K \chi_K \Cbb_K + \left[ \sum_K \chi_K \Cbb_K^{-1} \right]^{-1} \right)  \ .
\end{equation}
Note that assuming uniform stress or strain in all $\Omega_K$ in the domain $\Omega$  omits compatibility requirements that make the actual field dependent on differences in $\phib_K$ along interface (e.g.  orientation differences along inter-crystal boundaries (misorientation) \cite{kocks1998texture}).

\section{Datasets} \label{sec:data_sets}

To demonstrate the efficacy of the proposed architecture we will use a few exemplars: (a) heat conduction in polycrystals, (b) stress evolution in polycrystalline materials undergoing plastic deformation, and (c) stress evolution in viscoelastic composite materials.
The aspects of the data germane to machine learning are described in the following section and the remaining simulation details are given in \aref{app:data_models}.
With these exemplars we generate both 2D and 3D datasets.
For each exemplar only the apparent orientation or phase information and the discretization cell volumes are provided to the GCNN models.
The inclusion of cell volumes is motivated by the classical mixture formulas since the unstructured meshes have a distribution of element volumes.
The 3D datasets demonstrate the full generality of the neural network architectures described in \sref{sec:architecture} and test how they perform on samples with large discretizations, while the 2D datasets facilitate exploration through visualization.
Some of the datasets involve data pixelated on structured grids, which are treated as graphs by the graph convolutional neural networks,
These datasets enable direct comparison with pixel-based CNNs.

\subsection{Polycrystal heat conduction}  \label{sec:HF}

\begin{figure}[htb!]
\centering
\begin{subfigure}[b]{0.30\textwidth}
\centering
\includegraphics[width=1.00\textwidth]{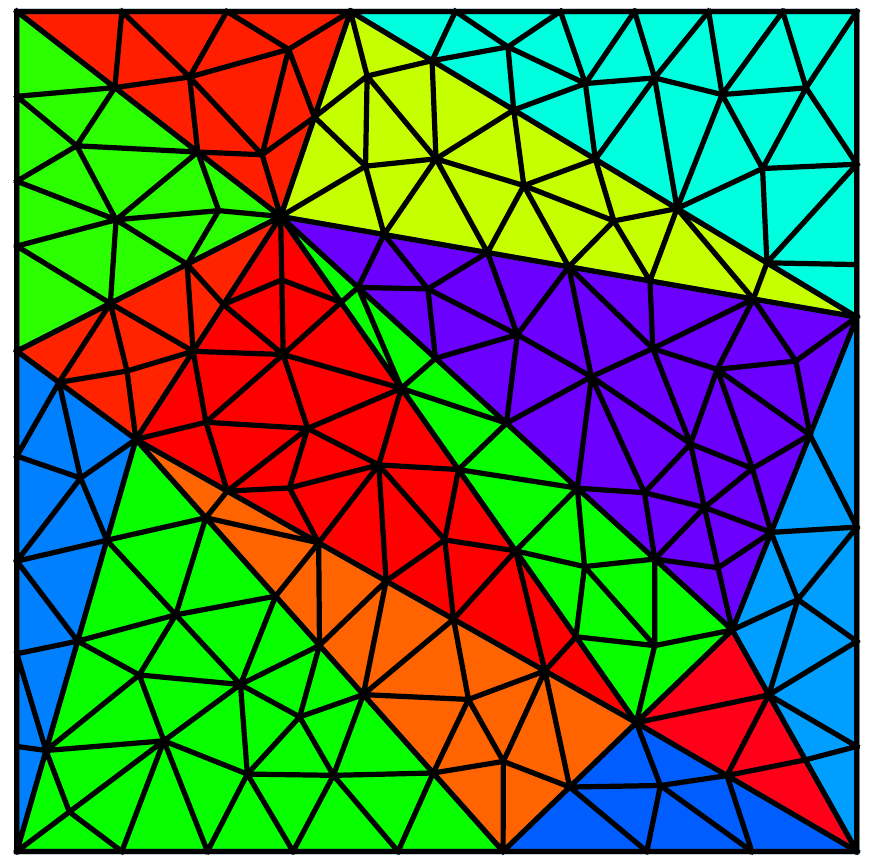}
\end{subfigure}
\begin{subfigure}[b]{0.30\textwidth}
\centering
\includegraphics[width=1.00\textwidth]{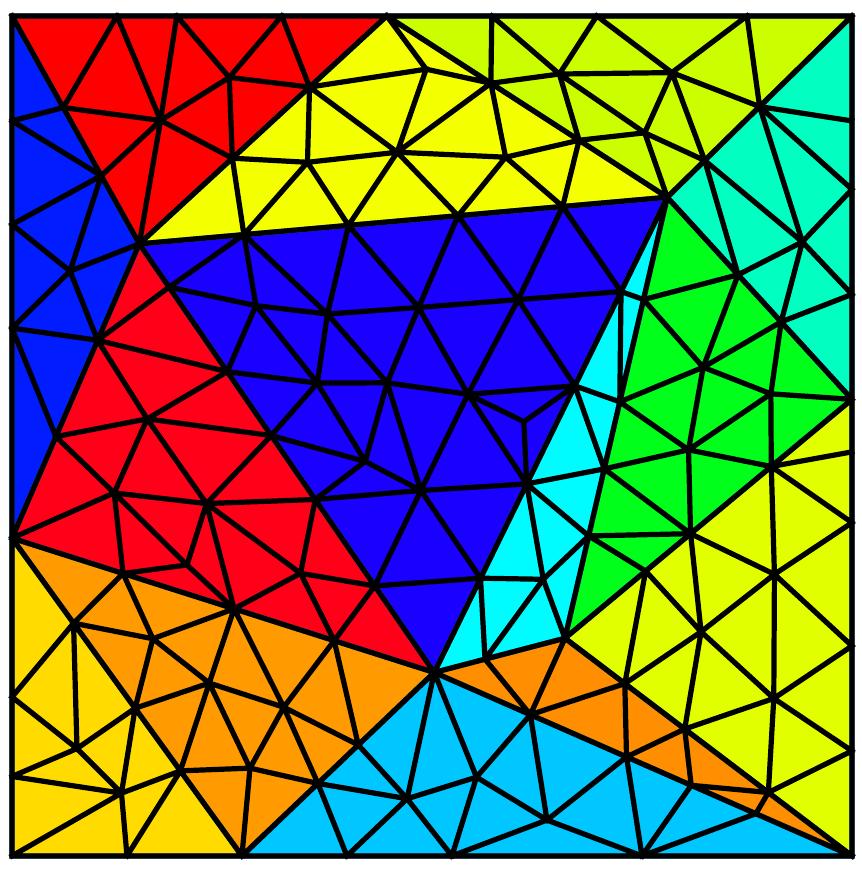}
\end{subfigure}
\begin{subfigure}[b]{0.30\textwidth}
\centering
\includegraphics[width=1.00\textwidth]{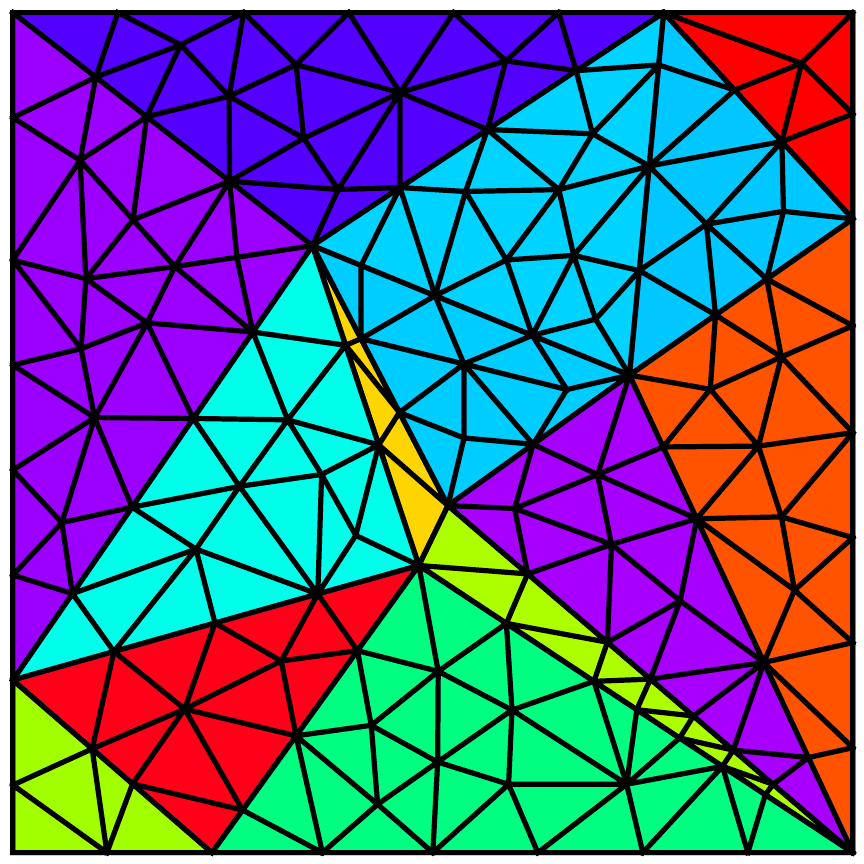}
\end{subfigure}

\begin{subfigure}[b]{0.30\textwidth}
\centering
\includegraphics[width=1.00\textwidth]{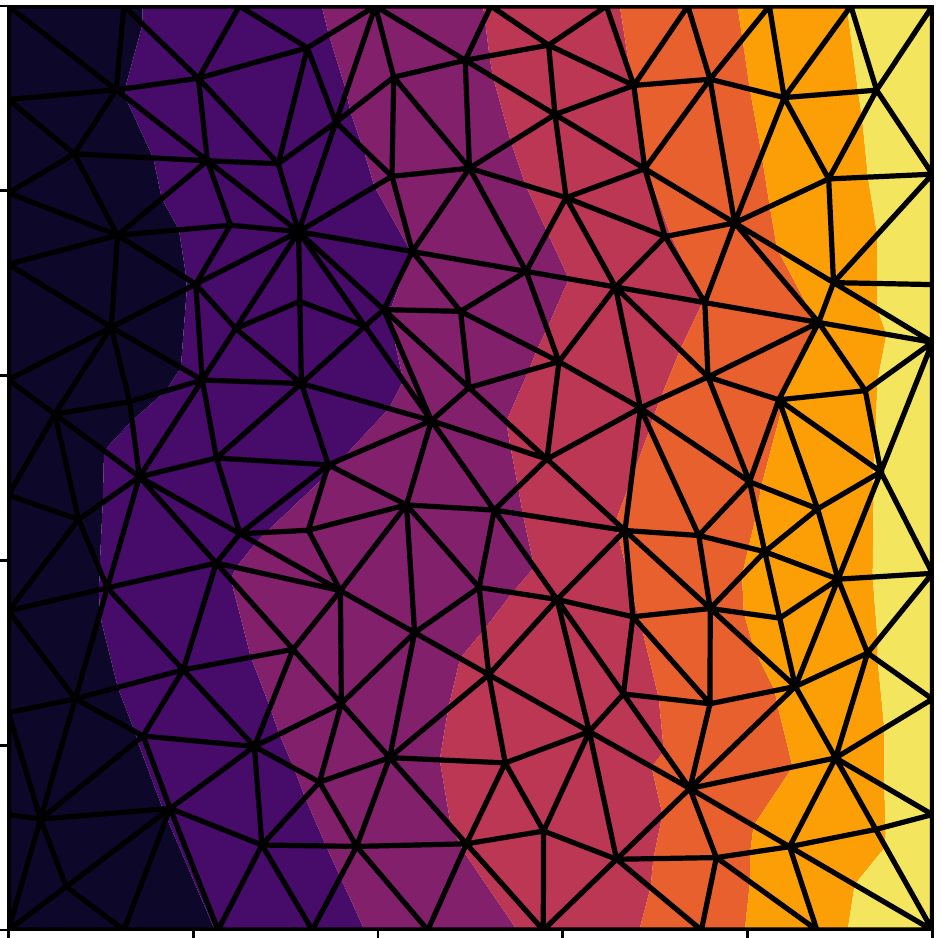}
\end{subfigure}
\begin{subfigure}[b]{0.30\textwidth}
\centering
\includegraphics[width=1.00\textwidth]{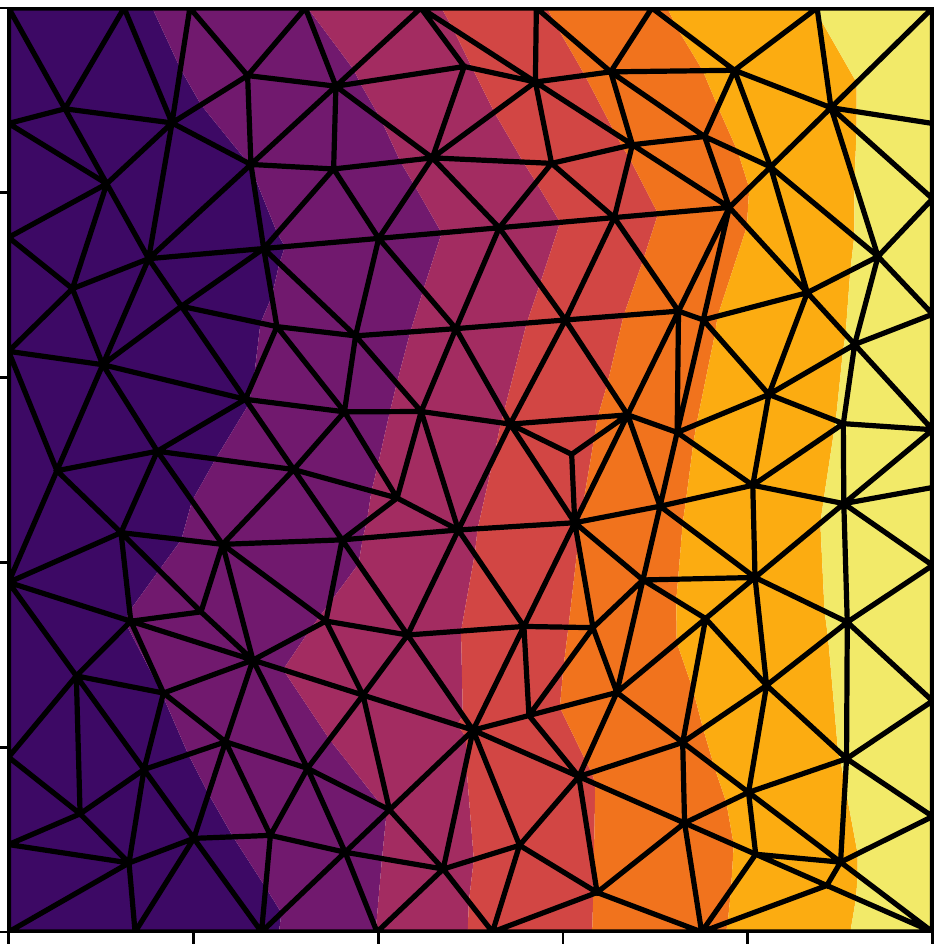}
\end{subfigure}
\begin{subfigure}[b]{0.30\textwidth}
\centering
\includegraphics[width=1.00\textwidth]{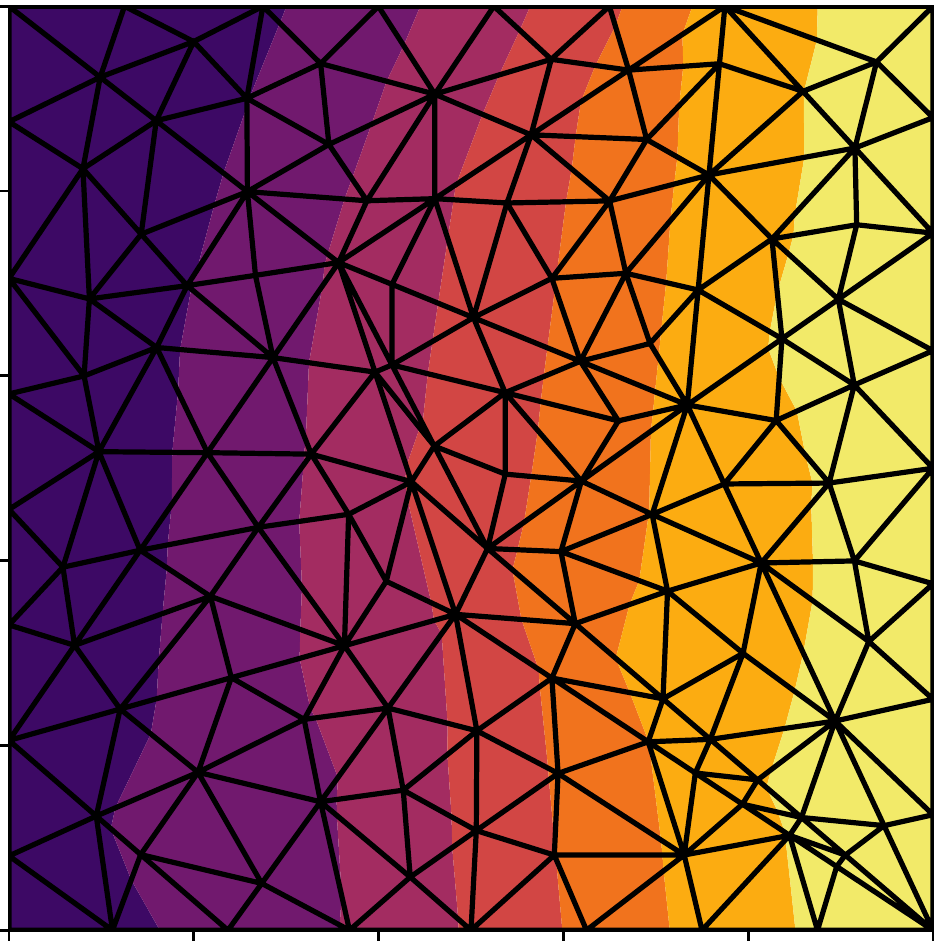}
\end{subfigure}
\caption{Thermal conductivity: representative realizations of orientation texture $\phi(\Xb)$ (upper row) and the corresponding temperature fields $\theta(\Xb)$ (lower row).
}
\label{fig:hf_realizations}
\end{figure}

The simplest dataset was generated with two-dimensional simulations of steady heat conduction through ersatz polycrystals represented on a triangular mesh.
The problem has the complexity that the conductivity tensor $\kappab$ in each crystal is orientation $\phib$ dependent; however, the governing partial differential equation  linear in the conductivity and the temperature fields.
\fref{fig:hf_realizations} shows a few of the 10,000 realizations generated, which which had 211 to 333 elements (mean 235.8) and 14 to 18 crystals (mean 15.4).Also apparent is deviation in the temperature fields from homogeneous gradients due to the anisotropy of the crystal components of the samples.

\begin{figure}[htb!]
\centering
\begin{subfigure}[b]{0.45\textwidth}
\centering
\includegraphics[width=1.00\textwidth]{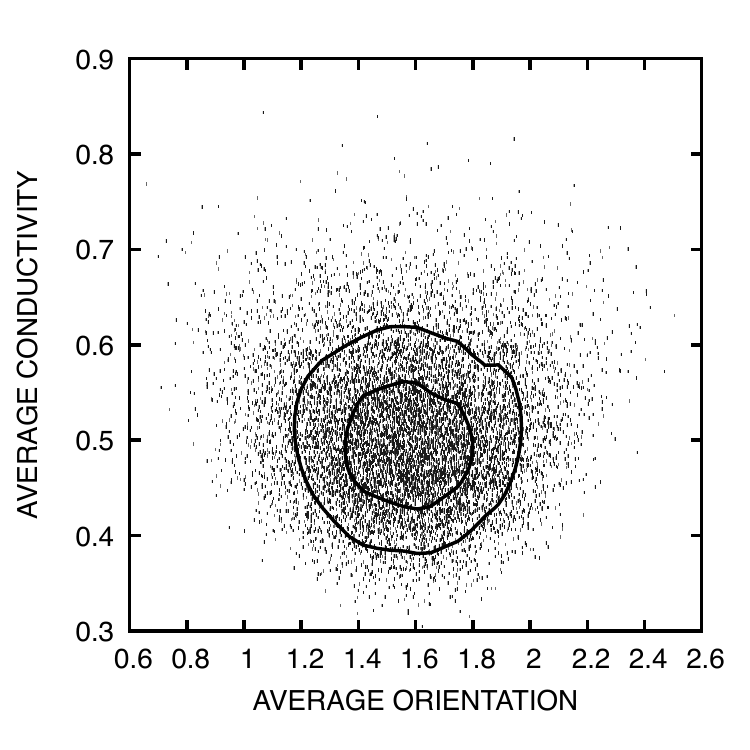}
\caption{conductivity-orientation correlation}
\end{subfigure}
\begin{subfigure}[b]{0.45\textwidth}
\centering
\includegraphics[width=1.00\textwidth]{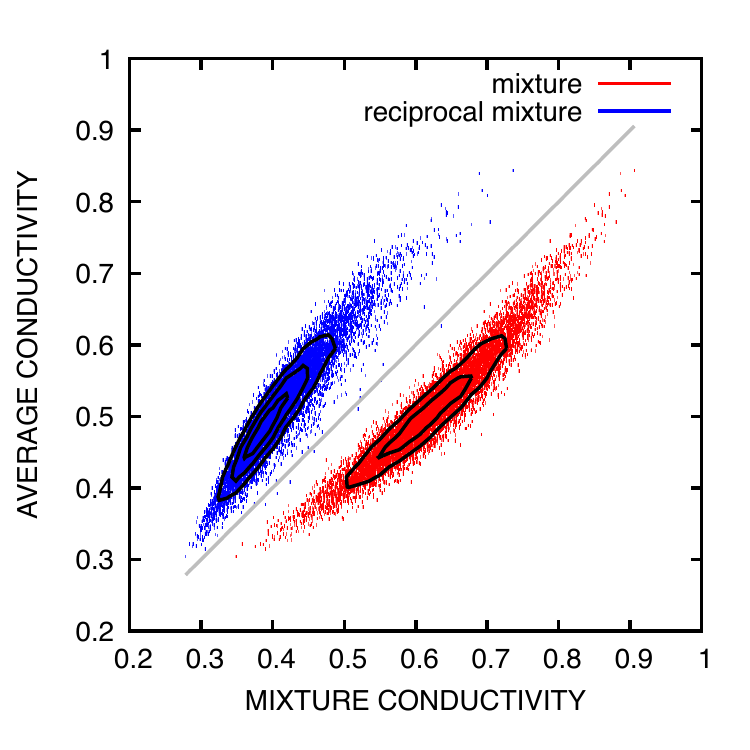}
\caption{conductivity-mixture correlation}
\end{subfigure}
\caption{Thermal conductivity: data correlations.
(a) average conductivity is essentially uncorrelated with the obvious feature of average correlation,
(b) average correlation is correlated with mixture and reciprocal mixture rules albeit with significant bias.
}
\label{fig:hf_correlations}
\end{figure}

As \fref{fig:hf_correlations}a shows, the volume averaged angle $\bar{\phi}$ for a realization is not particularly correlated with the output.
However if the crystal conductivities are known, mixture rules analogous to \eref{eq:Hill} for a homogeneous gradient
\begin{equation}
\bar{\kappa} = \sum_K \chi_K \kappa_K
\end{equation}
or homogeneous flux
\begin{equation}
\bar{\kappa} =  \left[ \sum_K \chi_K \kappa_K^{-1} \right]^{-1}
\end{equation}
give reasonably well-correlated estimates of the effective conductivity $\bar{\kappa}$.
These estimates are show in \fref{fig:hf_correlations}b for each of the realizations in the ensemble.
Clearly the mixture estimates are biased since the assumptions of homogeneous gradient or homogeneous flux are limiting cases.

\subsection{Polycrystals with crystal plasticity} \label{sec:CP}

\begin{figure}[htb!]
\centering
\begin{subfigure}[b]{0.30\textwidth}
\centering
\includegraphics[width=1.00\textwidth]{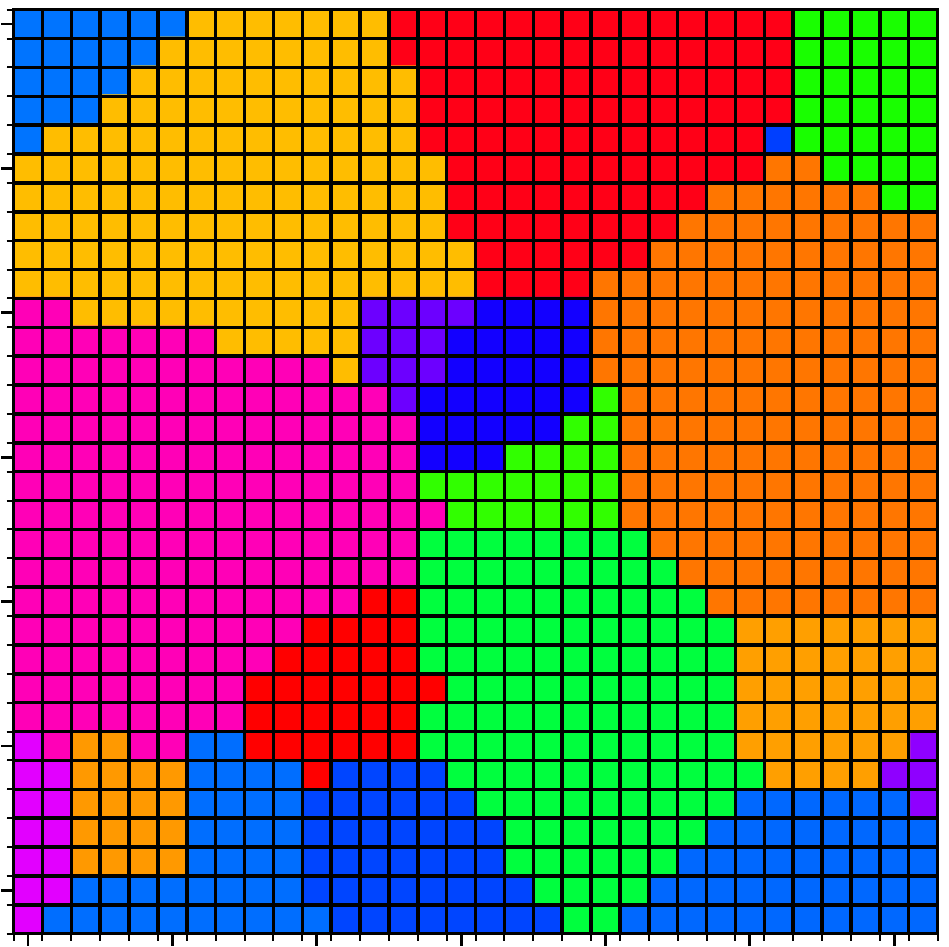}
\end{subfigure}
\begin{subfigure}[b]{0.30\textwidth}
\centering
\includegraphics[width=1.00\textwidth]{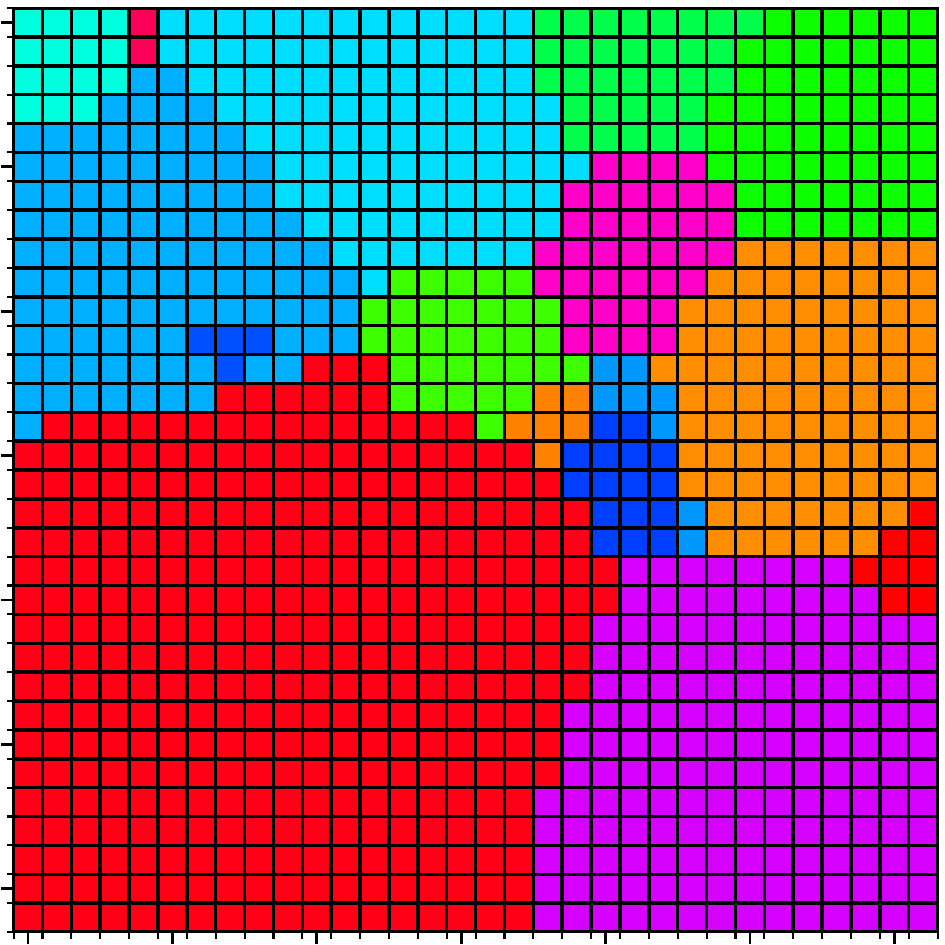}
\end{subfigure}
\begin{subfigure}[b]{0.30\textwidth}
\centering
\includegraphics[width=1.00\textwidth]{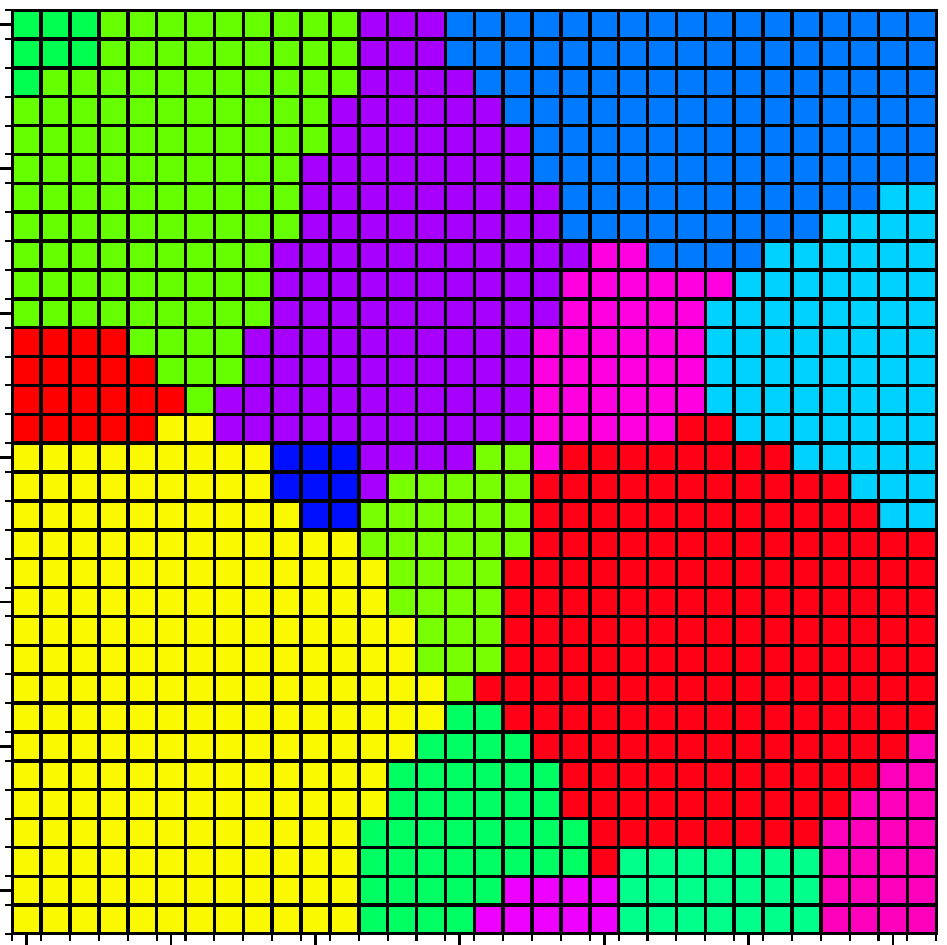}
\end{subfigure}

\begin{subfigure}[b]{0.30\textwidth}
\centering
\includegraphics[width=1.00\textwidth]{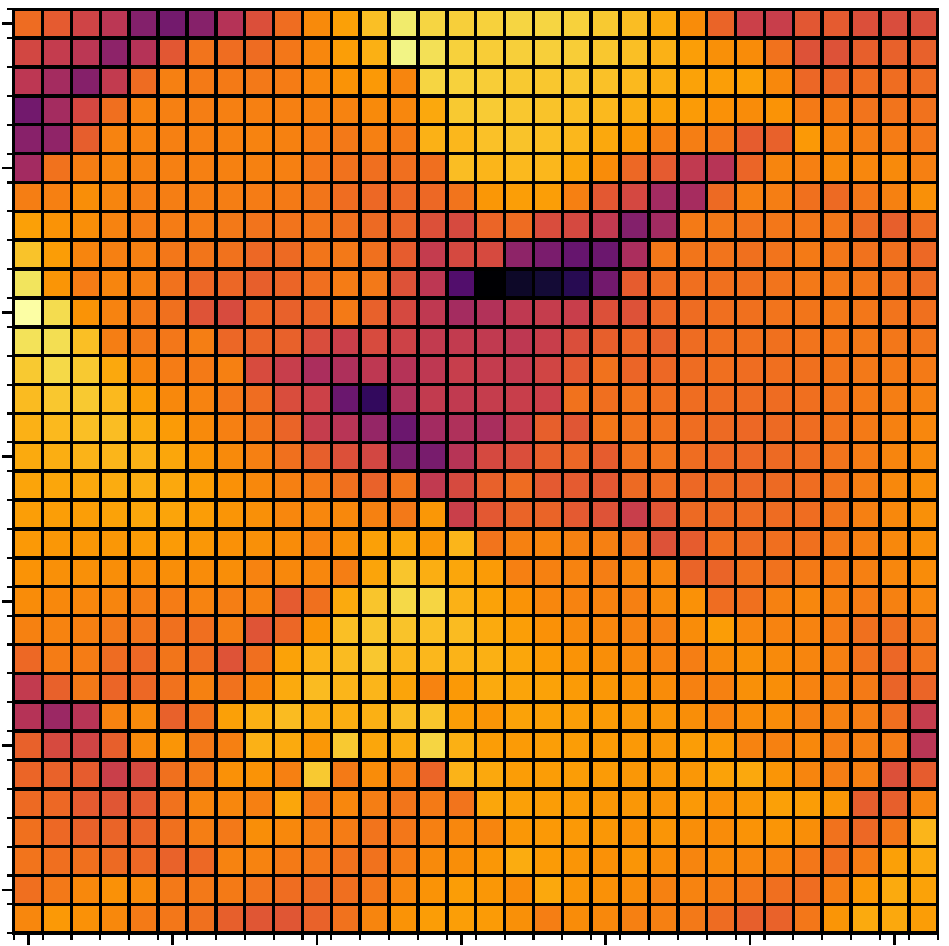}
\end{subfigure}
\begin{subfigure}[b]{0.30\textwidth}
\centering
\includegraphics[width=1.00\textwidth]{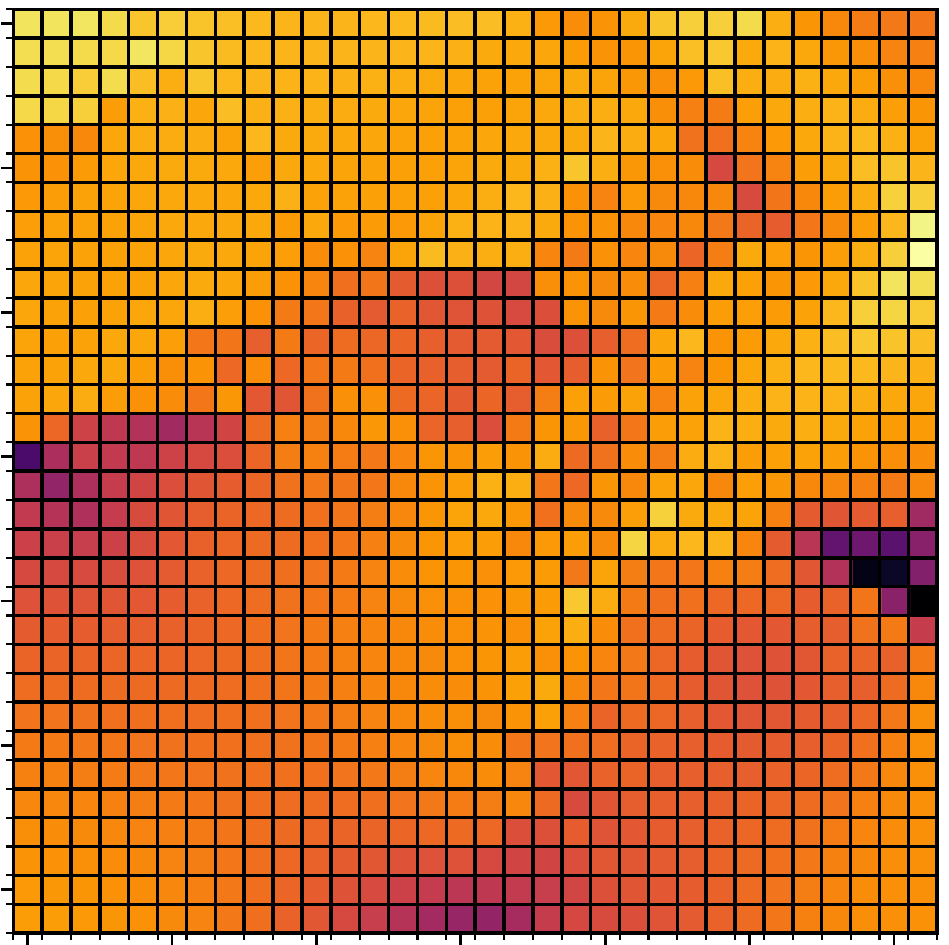}
\end{subfigure}
\begin{subfigure}[b]{0.30\textwidth}
\centering
\includegraphics[width=1.00\textwidth]{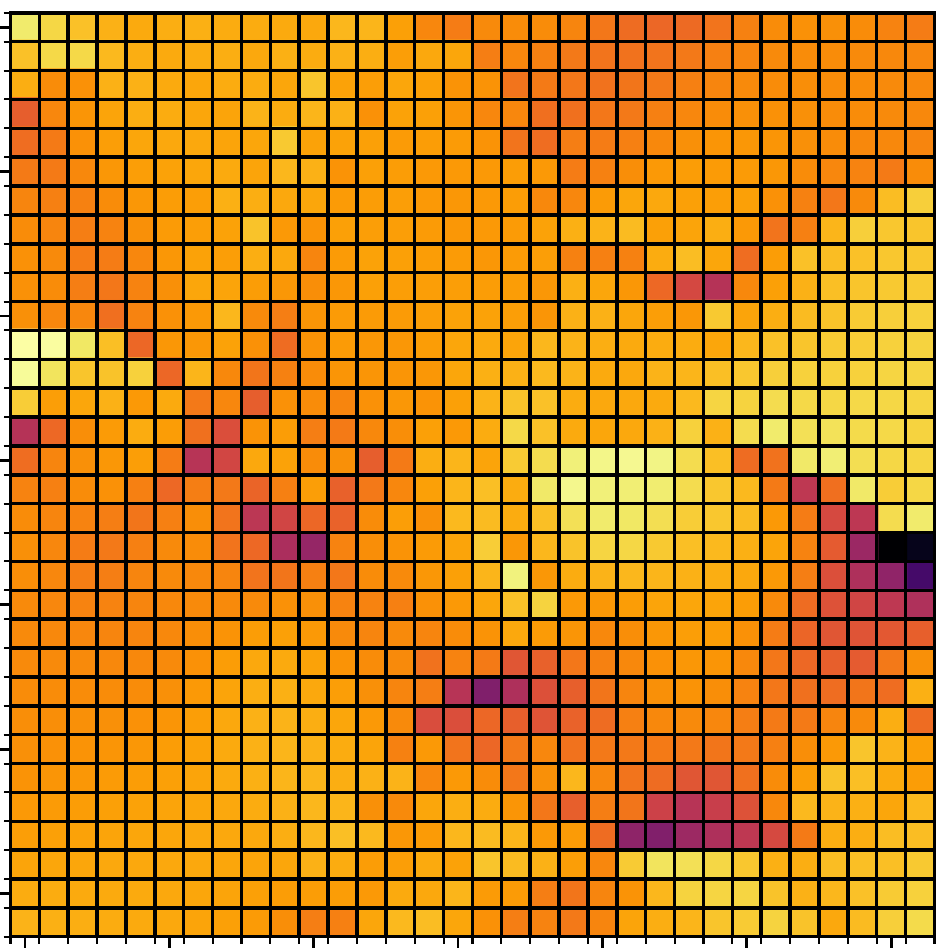}
\end{subfigure}
\caption{Stress: realizations of $\phi$ (upper row) and the corresponding $\sigma$ (lower row).
}
\label{fig:cp_realizations}
\end{figure}

Crystal plasticity (CP) is another common example where the homogenized response of a representative sample of grain structure characterized by the orientation angles $\phib(\Xb)$ is of interest.
As detailed in \aref{app:CP}, the response of the polycrystalline metal samples is complex and non-linear in the loading and properties.
\fref{fig:cp_realizations} shows a few of the 12,000 2D realizations on a 32$\times$32 grid and another 10,000 3D realizations on a 25$\times$25$\times$25 grid were simulated independently.
\fref{fig:cp_realizations} also illustrates the inhomogeneities in the stress response (at 0.3\% strain in tension), which are particularly marked at grain boundaries.
The 2D realizations had 5 to 30 grains (with a mean at 13.9) and the 3D realizations had 1 to 34 grains (with a mean at 13.0).

\begin{figure}[htb!]
\centering
\begin{subfigure}[b]{0.45\textwidth}
\centering
\includegraphics[width=1.00\textwidth]{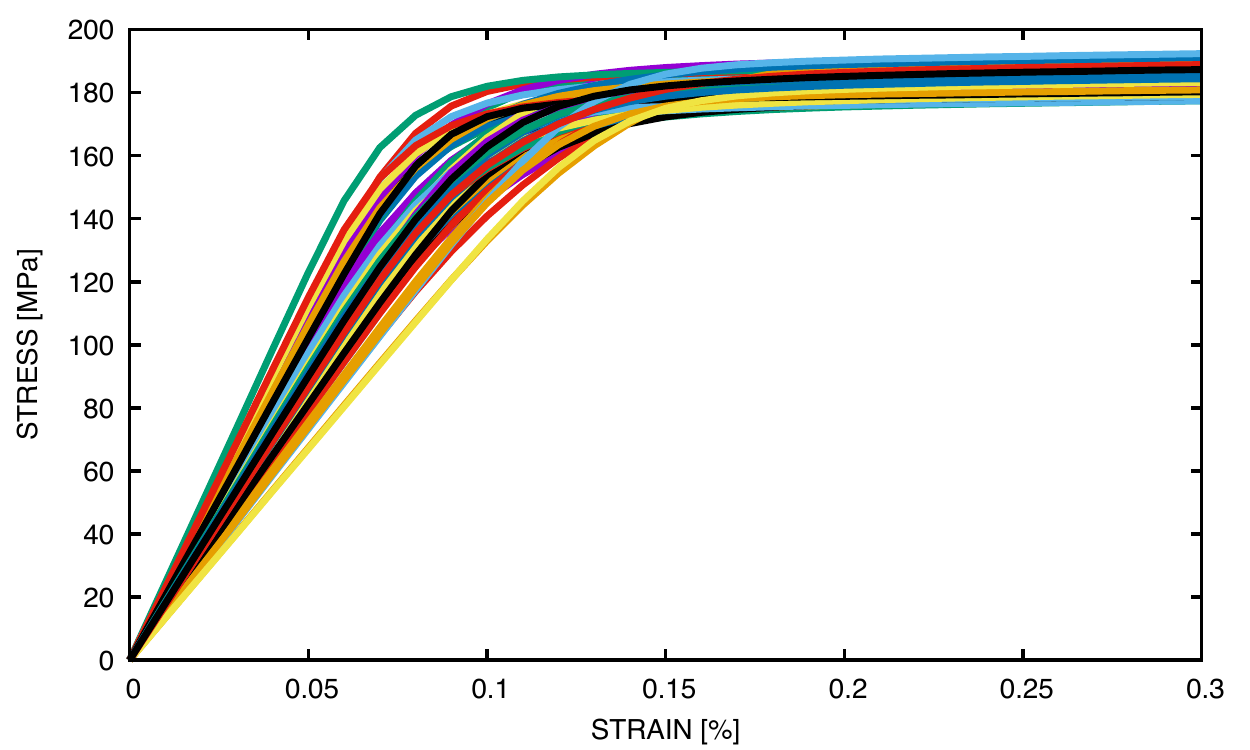}
\caption{stress evolution of an ensemble of samples (colors distinguish samples)}
\end{subfigure}
\begin{subfigure}[b]{0.45\textwidth}
\centering
\includegraphics[width=1.00\textwidth]{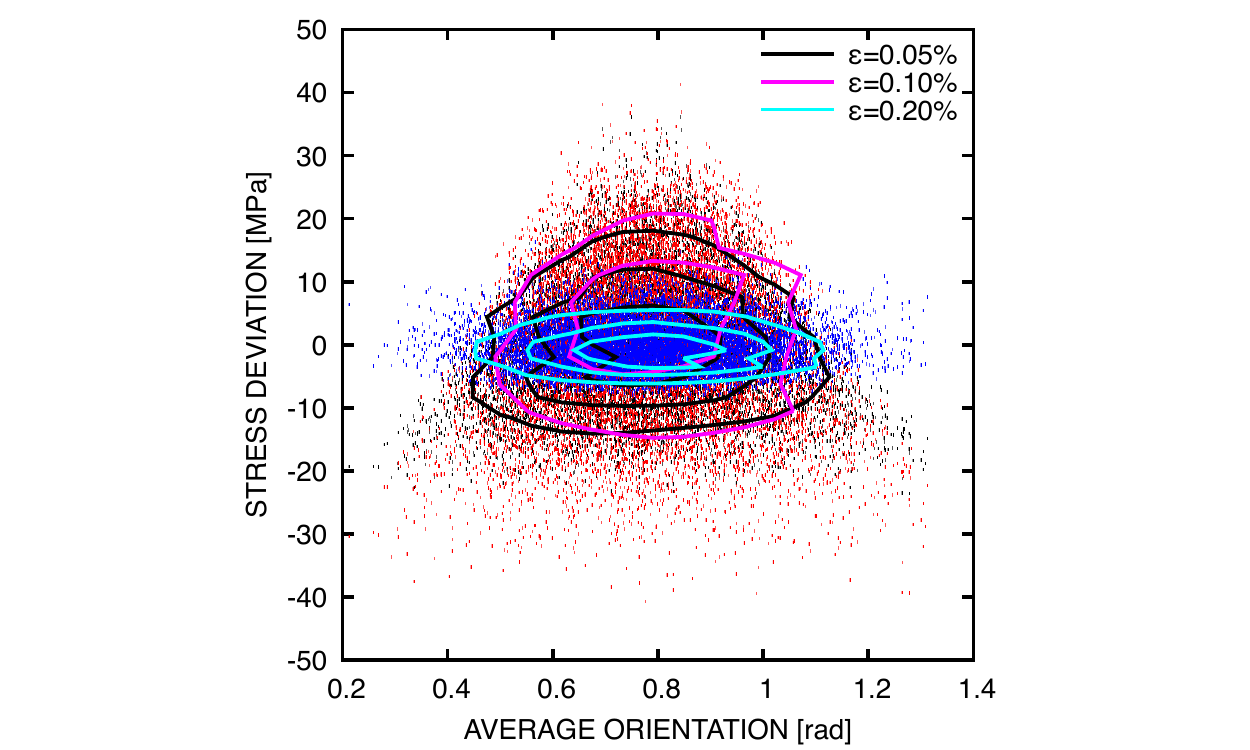}
\caption{stress-orientation correlations at various nominal strain levels (color contours correspond to density of same color point data)}
\end{subfigure}
\caption{Stress: realizations of $\sigma(t)$ and correlations of deviations $\sigma(t)$ from ensemble mean with average $\phi$.
}
\label{fig:cp_correlations}
\end{figure}

The resulting stress-strain response curves shown in \fref{fig:cp_correlations}a vary due to the particular texture $\phi(\Xb)$ of the realization.
Clearly the different crystal textures evoke different effective elastic slopes, yield points and (post-peak) flow stresses.
The correlation with the initial orientation decreases with strain and increasing plasticity as \fref{fig:cp_correlations}b shows.

\subsection{Composites of a viscoelastic matrix and stiff inclusions}  \label{sec:VE}

\begin{figure}[htb!]
\centering
\begin{subfigure}[b]{0.30\textwidth}
\centering
\includegraphics[width=1.00\textwidth]{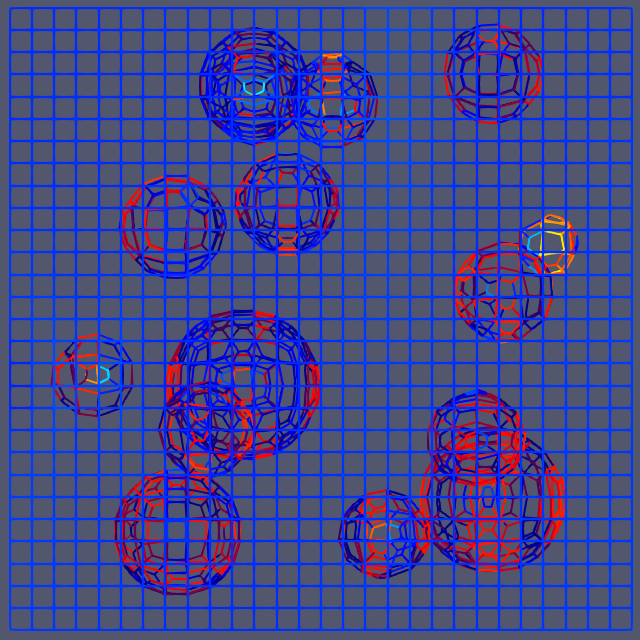}
\end{subfigure}
\begin{subfigure}[b]{0.30\textwidth}
\centering
\includegraphics[width=1.00\textwidth]{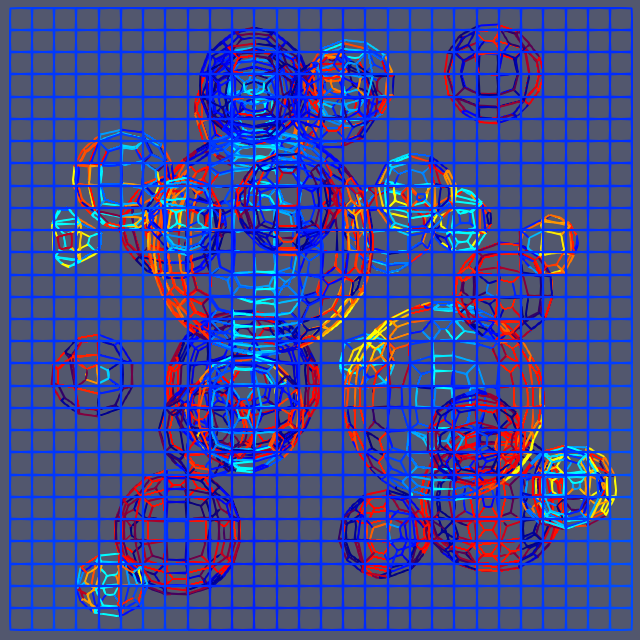}
\end{subfigure}
\begin{subfigure}[b]{0.30\textwidth}
\centering
\includegraphics[width=1.00\textwidth]{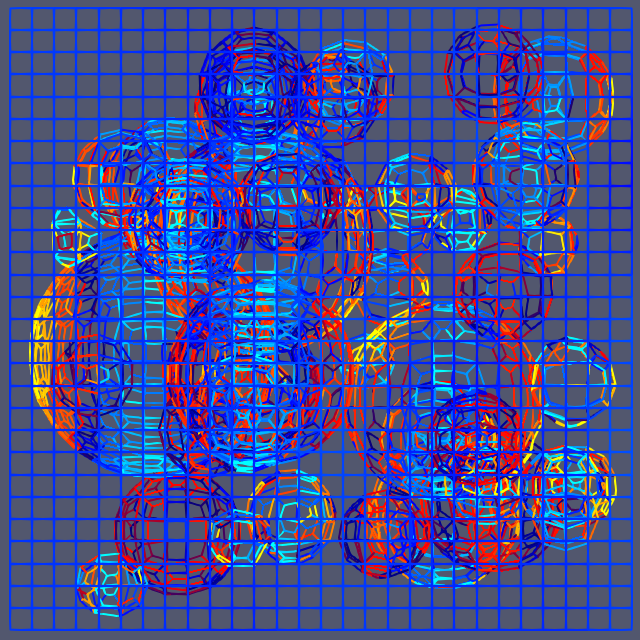}
\end{subfigure}

\caption{Stress $\sigma$ for viscoelastic matrix with stiff elastic inclusions for strain $\epsilon$=0.5 at time 83.33 $\mu$s.
}
\label{fig:pore_realizations}
\end{figure}

This exemplar studies another common microstructure: a composite material composed of a viscoelastic (polymer) matrix and stiff, spherical elastic (glass) inclusions.
Unlike the CP exemplar these realizations employ an unstructured mesh.
Another distinction from the previous exemplar is: for the matrix-inclusion composite most of the inclusions interact directly with the matrix (few interact directly with other inclusions), while in the CP exemplar grains have interfaces with a fairly uniform number of other grains.
\fref{fig:pore_realizations} illustrates the variety of inclusion sizes (polydispersity) and the stress concentrations at the inclusion-matrix boundaries (plotted in reference configuration at 50\% strain in tension).
For this dataset 3,000 3D realizations were generated with mesh sizes ranging from 23,152 to 28,892 elements (mean 25,430.9) with 16 to 24 inclusions (mean 17.3).

\begin{figure}[htb!]
\centering
\begin{subfigure}[b]{0.45\textwidth}
\centering
\includegraphics[width=1.00\textwidth]{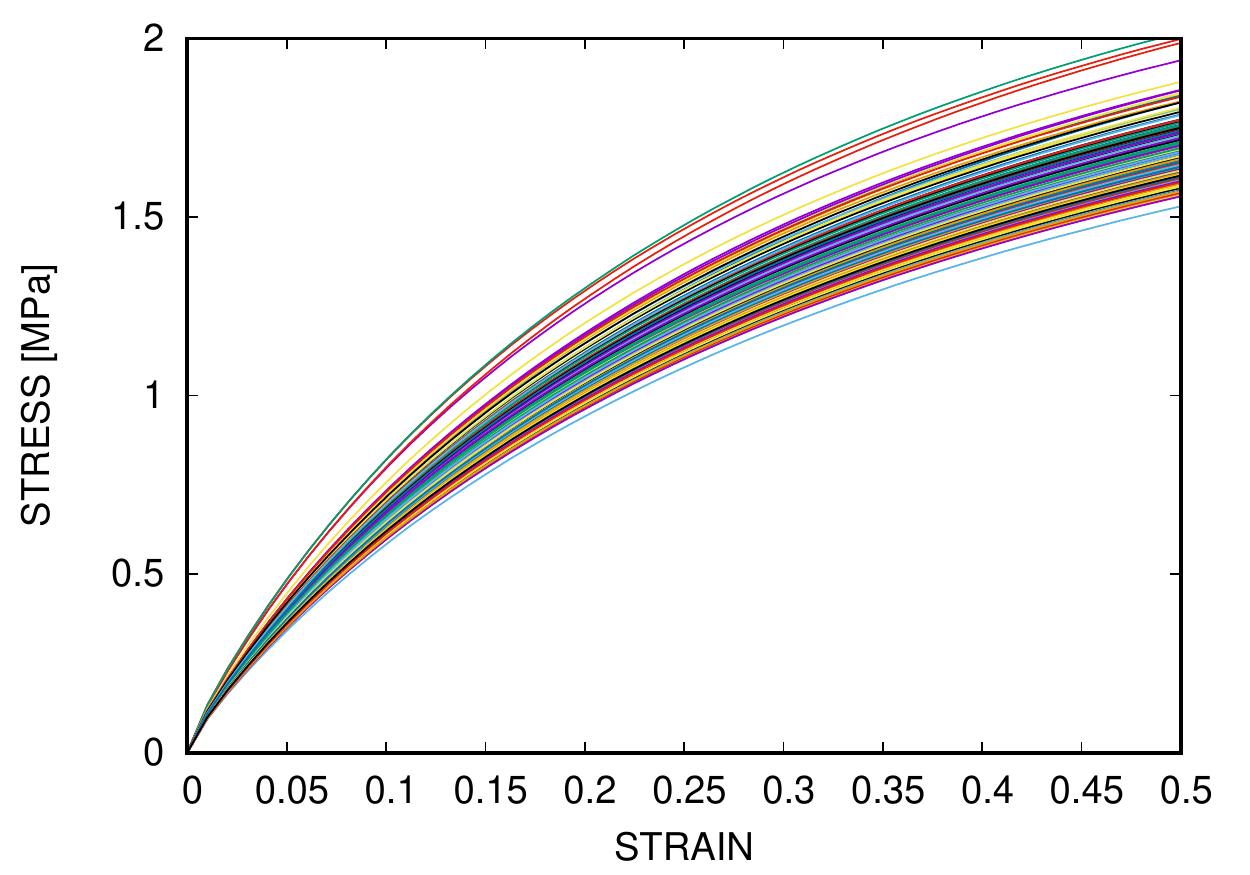}
\caption{stress evolution}
\end{subfigure}
\begin{subfigure}[b]{0.45\textwidth}
\centering
\includegraphics[width=1.00\textwidth]{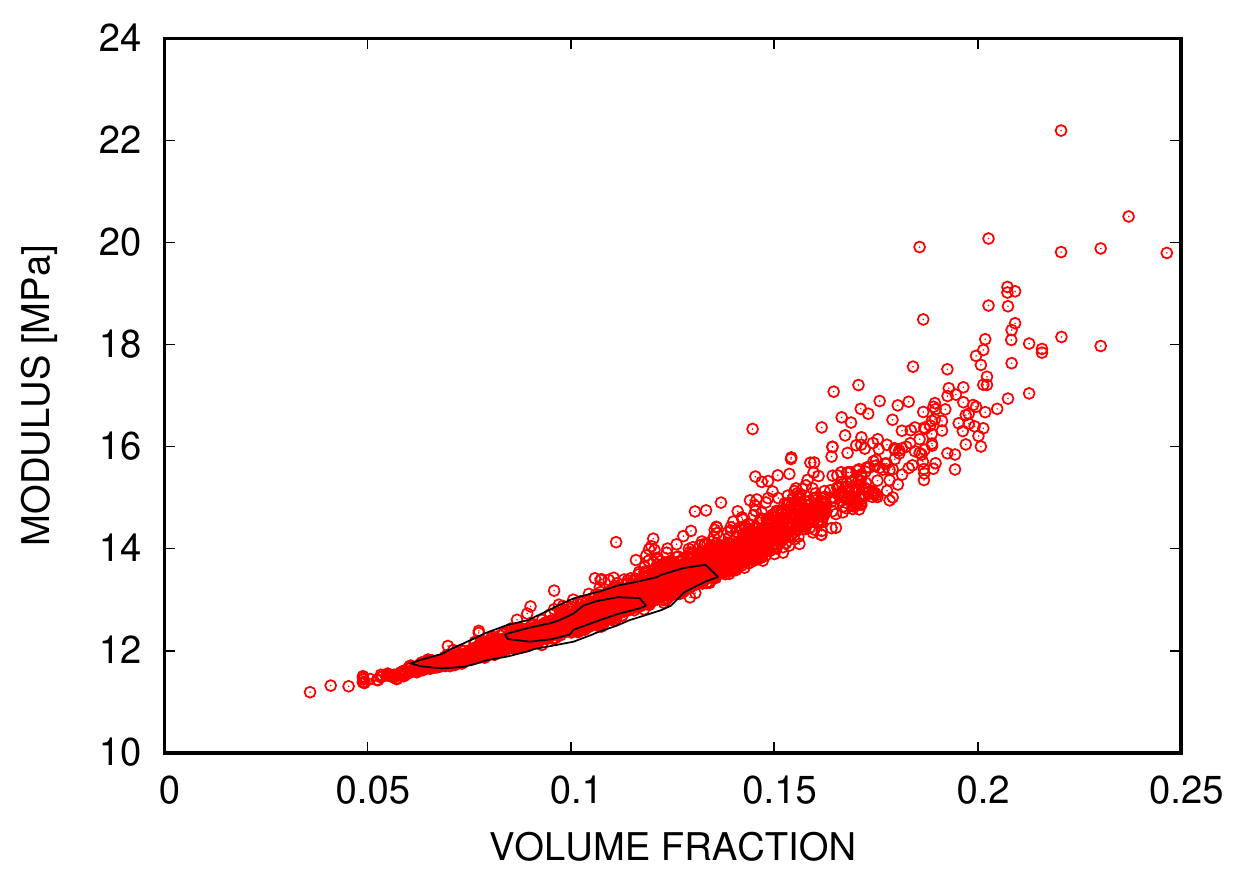}
\caption{volume fraction-modulus correlations}
\end{subfigure}
\caption{Stress: realizations of $\sigma(t)$ (where colors distinguish realizations) and correlations of the initial modulus $\partial_\epsilon \sigma(t=0)$ with the average volume fraction $\phi$.
}
\label{fig:pore_correlations}
\end{figure}

As \fref{fig:pore_correlations}a shows, the stress responses largely self-similar; however, the initial modulus plotted in \fref{fig:pore_correlations}b (nor the average stress) is not perfectly correlated with average porosity as it would be in a simple mixture.
In this case $\phi(\Xb) = 0$ for the matrix and $\phi(\Xb) = 1$ for an inclusion, i.e. the properties of the constituents are not included in the training data.

\section{Graph neural network architecture} \label{sec:architecture}

In this section we describe the proposed multi-level graph neural network for deep learning of homogenized response using segmented images of initial microstructure.
The network we propose is deep to address the question of what features need to be associated with the nodes of the cluster-level graph in order to create an accurate model.
Unlike pixelized images used in CNNs, graph representations of clustered image data typically can not be reconstructed into images and therefore are incomplete embeddings of the information in the input images.
This is even true of graphs that embed distances in their edge features or adjacency matrices; and, clearly, the usual binary adjacency induces a different sense of distant between nodes than in the original image.
Some features beyond the image values in the subdomains are obvious, such as the cell volumes used in mixture homogenization formulae; whereas, others are not and discovering the hidden features will be the task of the proposed deep neural network.
In the proposed architecture we also seek to take advantage of multiple physical scales to interactions, e.g. short-range interactions at the level of the discretization and longer range at the level of the clusters.
By combining information at multiple levels in a deep manner we expect to produce a more accurate model.

In our previous work~\cite{frankel2022mesh}, we showed that GCNNs based on the native discretization of the image data, whether on unstructured meshes or structured grids, provided an effective representation for homogenization tasks.
This approach can become prohibitive for very fine discretizations since the adjacency matrices and node data scale with the size of the discretization.
Unlike graph clustering solely based on node adjacency~\cite{dhillon2007weighted} independent of data on graph, here the initial data $\phib$ clearly informs the clustering of the adjacency derived from the computational mesh.
So we take an approach similar to the soft assignment GNNs discussed in \sref{sec:intro}~\cite{ying2018hierarchical,bianchi2019graph}; however, we provide a cluster assignment matrix of the mesh entities based on SVE subdomains as data.

\begin{figure}[htb!]
\centering
\begin{subfigure}[b]{0.85\textwidth}
\includegraphics[width=0.9\textwidth]{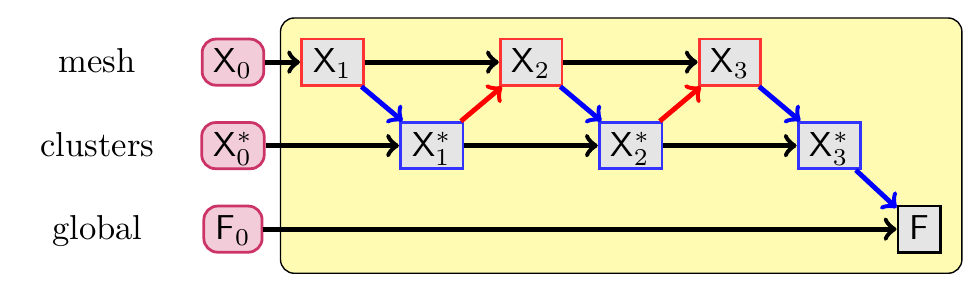}
\end{subfigure}
\caption{Generalized multi-level architecture for the convolutional unit (yellow in \fref{fig:architecture}.)
Levels:
(1) ``mesh'' at the scale of the discretization, with input node data $\Xs_0$ being phase or orientation, for example;
(2) ``clusters'' at the scale of the logical units/phases, with input node data $\Xs^*_0$ being volumes, for example; and
(3) ``global'' at the system/sample scale, with input data $\Fs_0$ being volume fraction, for example.
Connections:
(1) horizontal black arrows: convolution (or direct pass-through);
(2) downward blue arrows: reduction via $\Ss \Xs$; and
(3) upward red arrows: prolongation via $\Ss^T \Xs^*$.
Multiple arrows arriving at a state represents concatenation.
}
\label{fig:general_architecture}
\end{figure}

\begin{figure}[htb!]
\centering
\begin{subfigure}[b]{0.45\textwidth}
\includegraphics[width=1.0\textwidth]{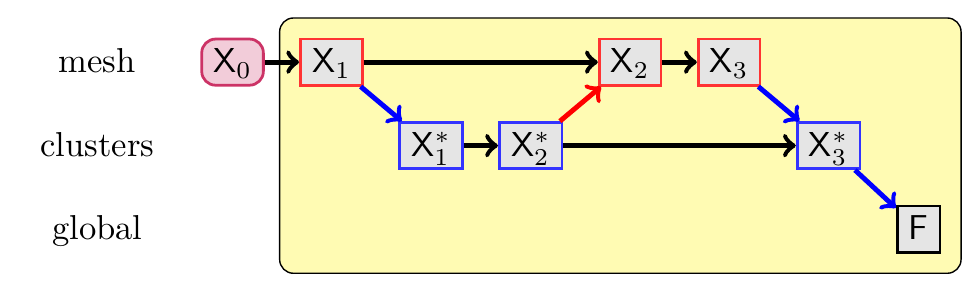}
\caption{Vee (vGCNN)}
\end{subfigure}
\begin{subfigure}[b]{0.45\textwidth}
\includegraphics[width=1.0\textwidth]{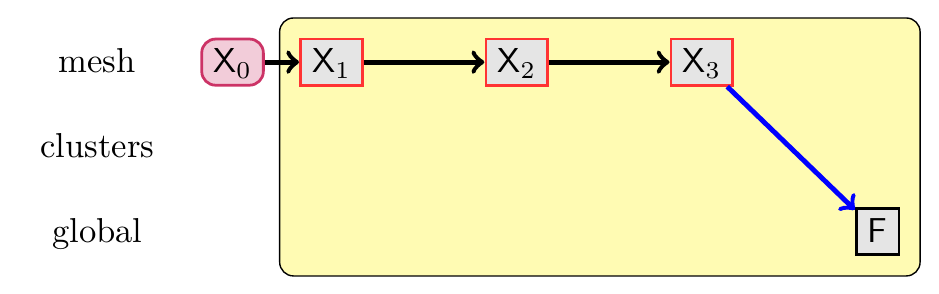}
\caption{Full (fGCNN)}
\end{subfigure}

\begin{subfigure}[b]{0.45\textwidth}
\includegraphics[width=1.0\textwidth]{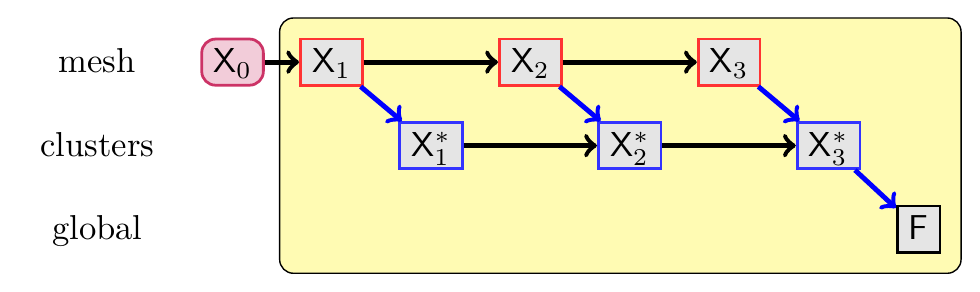}
\caption{Down (rGCNN)}
\end{subfigure}
\begin{subfigure}[b]{0.45\textwidth}
\includegraphics[width=1.0\textwidth]{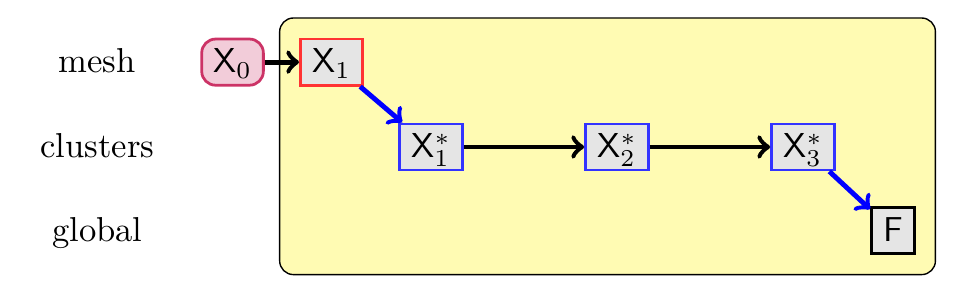}
\caption{Reduced (rGCNN)}
\end{subfigure}
\caption{Multilevel GCNN architecture variants
}
\label{fig:architecture_variants}
\end{figure}

Restricting our multi-level conception of the graph appropriate for microstructure homogenization problem to three levels: discretization, cluster, and sample, a general framework is suggested by \fref{fig:general_architecture}.
In this schematic inputs on the left are transmitted to outputs on right through a sequence of operations.
Inputs can be known features at the discretization/mesh (e.g. element volumes), cluster (e.g. phases), or sample/global level (e.g. volume fraction).
Convolution (black horizontal arrows) with trainable filters extract informative features correlated with output of interest and maintains the level of the nodal data.
The reduction (downward blue arrows) and prolongation (upward red arrows) operations exchange information at levels.
\fref{fig:architecture_variants} illustrates four specific variants of the framework.
The {\it vee} GCNN (vGCNN, \fref{fig:architecture_variants}a) is an example where information can flow cluster to discretization and vice versa.
Whereas, in the {\it down} variant (dGCNN, \fref{fig:architecture_variants}c) only finer to coarser information flow is allowed (discretization to cluster and cluster to global)
The {\it reduced} GCNN (rGCNN, \fref{fig:architecture_variants}d) is a simplified version where only one convolution at discretization is prescribed.
(The reasons for this design discussed in more detail later in this section.)
For comparison, the GCNN operating at the full discretization level \cite{frankel2022mesh} (fGCNN, \fref{fig:architecture_variants}b) is also shown.

\begin{figure}[htb!]
\centering
\includegraphics[width=0.9\textwidth]{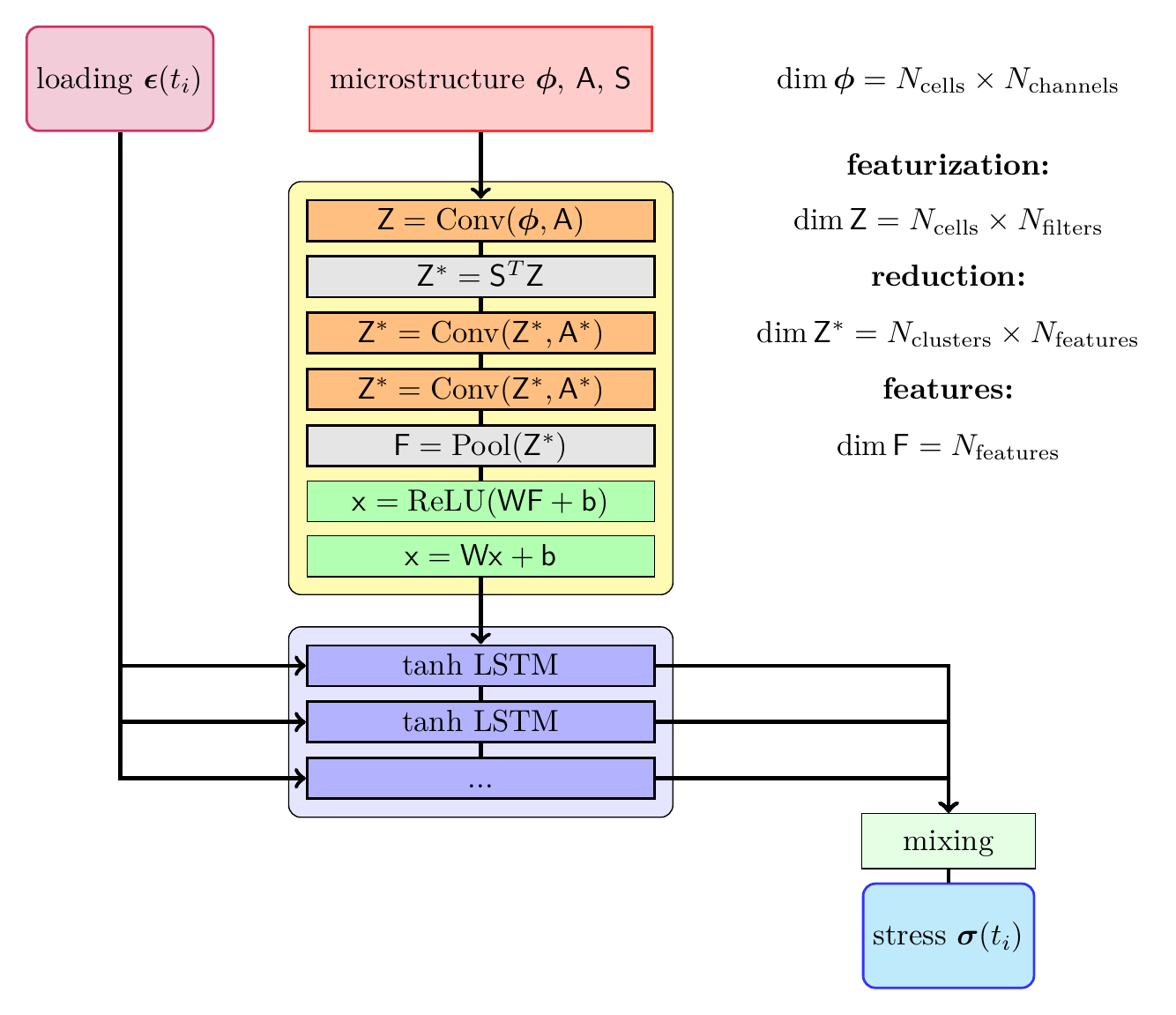}
\caption{Reduced GCNN architecture based on known (input-based) segmentation $\Ss$ and a mesh-based adjacency $\As$.
}
\label{fig:architecture}
\end{figure}

Taking the rGCNN as an example, a more detailed schematic is shown in \fref{fig:architecture}.
The inputs for the architecture, are the discretization-level node feature matrix $\phib$ of selected features on image/computational cells, the adjacency matrix $\As$ formed from the topology of the native discretizaton (mesh), and a clustering assignment matrix $\Ss$ determined by the SVE subdomain segmentation.
Each input is unique to the particular SVE sample and consists, for example, of the obvious features from imaging: phase, orientation and from the discretization: volume of the pixel or element.
The adjacency matrix $\As$ is constructed directly from the mesh connectivity and it is a symmetric binary matrix $N_\text{cells} \times N_\text{cells}$ with ones for row $i$ and column $j$ where cells $i$ and $j$ are considered neighbors and zeros elsewhere:
\begin{equation} \label{eq:A}
A_{ij} = \begin{cases}
1 & \text{if} \ \text{nodes} \ i,j \ \text{are neighbors} \\
0 & \text{else}
\end{cases}
\end{equation}
The (known) data segmentation as encoded by a $N_\text{cells} \times N_\text{clusters}$ matrix $\Ss$ which maps elements to clusters in {\it one-hot} type encoding such that any row of $\Ss$ has only one ``1'' and the rest zeroes,
\begin{equation} \label{eq:S}
S_{Ki} = \begin{cases}
1 & \text{if} \ \text{node} \ i \ \text{is in cluster} \ \Omega_K \\
0 & \text{else}
\end{cases}
\end{equation}
where the columns are assigned to regions $\Omega_K$ in no particular order.
The matrix $\Ss$ acts as a {\it restriction} by mapping a vector in the full/mesh space to the reduced and $\Ss^T$ acts as a {\it prolongation} by a vector in the reduced/cluster space to the full.
For instance, the orientation on the full graph is given by
\begin{equation}
\phib = \Ss^T \phib^*
\end{equation}
where $\phib^*$ is a vector of orientation assignments per region $\Omega_K$;
and, conversely, the vector of cluster volumes is given by
\begin{equation}
\Vs^* = \Ss \Vs
\end{equation}
where $\Vs$ is a vector of element volumes.

As in our previous work~\cite{frankel2022mesh,jones2021neural}, the core of the model is based on graph convolution which can be expressed as
\begin{equation} \label{eq:gconv}
\Zs' = \operatorname{Conv}(\Zs,\As) \equiv a(\Ws \Zs + \bs)
\end{equation}
where $a$ is a non-linear activation function applied element-wise, $\Ws$ is a trainable weight matrix and $\bs$ is a trainable bias vector.
Here we use ReLU activations based on previous work~\cite{frankel2022mesh}.
The weight matrix $\Ws$ is composed of the sum of products of $w_K$ trainable scalars and $\As_K$ pre-defined adjacency matrices:
\begin{equation} \label{eq:wconv}
\Ws = \sum_K w_K \As_K \ ,
\end{equation}
for instance $\As_0$ describes self interactions and $\As_1$ describes first nearest neighbors.
As in \cref{frankel2022mesh} we use a 2 weight GCNN (independent self and neighbor weights), as opposed the commonly employed Kipf and Welling convolution that makes the self weight dependent on the neighbor weight \cite{kipf2016semi}.

Clustering, as guided pooling, reduces the full adjacency $\As$ to the adjacency on the reduced graph
\begin{equation}
\As^* = \Ss^T \As \Ss
\end{equation}
subject to subsequent normalization.
The on-diagonal components scale  like $\As^*_{II} = \sum_{i,j\in\Omega_I, i\ne j} \approx \bar{n} N_I 1 \propto V_I$ and the off-diagonal like $\As^*_{IJ} = \sum_{i\in\Omega_I,j\in\Omega_J} 1 \approx 2 (\bar{n}/2) N_{IJ} \propto A_{IJ} $ where $\bar{n}$ is the average number of neighbors, $V_I$ is the volume of the cluster, and $A_{IJ}$ is the surface area between clusters $I$ and $J$.
The reduced graph adjacency clearly encodes relevant physical information about the SVE.
The discrepancies in the different scaling of the diagonal and off-diagonal entries are handled by the two independent trainable weights.

The assignment matrix $\Ss$ also plays a role in generating the node features $\Zs^*$ on the reduced graph but it can be ineffective to merely restrict the input data $\Zs^* = \Ss^T \Zs$ (see \cref{frankel2022mesh} and \sref{sec:variants}), instead they are attained by a trainable convolution
\begin{equation} \label{eq:featurization_convolution}
\Zs^* = \Ss^T \Conv(\Zs,\As)
\end{equation}
as in \cref{ying2018hierarchical}.
This layer (or layers) does the task of featurization, \ie determining the combinations of data from the source image that are needed to create an effective model of the output.
The number of filters of this layer $N_\text{filters}$ limit the number of features $N_\text{features}$ of the node data $\Zs^*$ on the reduced graph; however, as we will demonstrate in \sref{sec:hidden} this is not a significant hindrance since selecting $N_\text{filters}$ greater than the necessary number of features leads to a model with similar performance to a tuned one.
We expect $N_\text{filters} \ge N_\text{channels}$ to encode information lost in the graph embedding.

Since one of the objectives is to achieve comparable performance to a GCNN operating solely on graph representations of the mesh discretization~\cite{frankel2022mesh}, we examine the post-featurization convolution (refer to \fref{fig:architecture}):
\begin{equation}
\operatorname{Conv}(\Zs^*,\As^*)
= \Ws^* \Zs^* + \bs^*
= \Ss \Ws \Ss^T \Ss \Zs + \bs^*
= \Ss \left[  \Ws \Ss^T \Ss \Zs + \bs \right] \ ,
\end{equation}
which would be the reduction of the convolution on the full graph $\Ws \Zs + \bs$ if $\Ss^T \Ss \Zs = \Zs$.
Here, $\Ws$ and $\bs$ the weights and biases of the convolution on the full graph, and $\Ws^*$ and $\bs^*$ the weights and biases on the corresponding reduced graph.
The assignment matrix is such that
\begin{equation}
\sum_k S_{Ik} S_{Jk} = \sum_k S_{Ik}^2 \delta_{IJ} = N_I \delta_{IJ} \ ,
\end{equation}
where $N_I$ is the number of cells in cluster $I$, so we can normalize this product by the cluster sizes.
This normalization effects a cluster-wise average.
In analog to global pooling operations (sum, average, min, max) other cluster level pooling can be effected with difference normalizations.
We found that the reduction operations layer that process the inputs $\phib$ it was best to use sum-like reductions that result in a total for extensive inputs like volumes, and average reductions for intensive variables like orientation.
Note the equivalence between the convolution on the full graph and the reduced graph only holds when the node data $\Zs$ is uniform in the clusters, which is true for the input $\Zs$ but not after subsequent convolutions.
In this work we only employ one featurizing convolution \eref{eq:featurization_convolution} with nearest neighbor adjacencies; however, a stack of 1-hop convolutions or a $k$-hop convolutions could be employed.

The remainder of the network operates on the reduced graph and associated features and culminates in a global pooling to create system-level features.
Our previous work \cite{frankel2022mesh} and others \cref{bianchi2020spectral} has indicated that a skip connection with an  independent self weight is beneficial, so we use graph convolutions with 2 trainable $w_K$ as in \cref{frankel2022mesh}.
Since the number of output features is determined by the number of filters in this convolutional stack, we denote them as $N_\text{features}$.
The final global pooling/average shown in \fref{fig:architecture}
\begin{equation}
\Pool(\Zs) = \frac{1}{N_\text{cells}} \sum_i Z_i
\end{equation}
accounts for different number of clusters differ across samples and provides global scalar features $F_a$ that are mixed by a linear dense layer to be used as input to an MLP for property prediction, for example effective conductivity \sref{sec:HF}, or a recurrent neural network for evolution prediction, as with the crystal plasticity exemplary \sref{sec:CP}.

The architecture presented above was implemented via TensorFlow \cite{tensorflow} and Spektral \cite{spektral}.

\section{Interpretation methods} \label{sec:interpret}

By design in the proposed architecture features $F_a$ are equivalent to the average of the final filter outputs, refer to \fref{fig:architecture}, in a loose analogy with traditional homogenization estimators discussed in \sref{sec:problem}.
We can correlate these features $F_a$ with the physical outputs of the network, \eg $\sigmab(t)$
\begin{equation}
\rho_{F_a, \sigma} = \frac{\Vbb[ F_a, \sigma ] }
{\sqrt{\Vbb[F_a] \Vbb[\sigma]}}
\end{equation}
to determine what regimes are most active.
Here, $\Vbb[x]$ is the variance of $x$ and $\Vbb[x,y]$ is the covariance of $x$ and $y$.
Filter activation can be assessed by the statistics of whether the filter of a particular convolutional layer produces non-zero output over the ensemble of input images $\phib(\Xb)$.
The activation may change over different regimes in the response $\bar{\sigma}(\epsilon)$, which can indicate specialization to, for example, the elastic response or plastic flow.
Simple filter visualization pairing the output image given an input image can also be illuminating since it can show what regions activate the filter.
(In the case of a GCNN the map from the graph output to an image is based on node assignments to source mesh.)
Filter maps of the input $\phi(\Xb)$ to the filter output of the last convolutional layer $\Xs_{n}(i,a)$, corresponding to the global pooled feature $F_a$ can also be constructed and visualized with density plots.
Note that the redundancy that is common in NNs confounds interpretation, so to aid interpretability with the architecture shown in \fref{fig:architecture} we simplify the previously proposed network architecture \cite{frankel2022mesh} by omitting dense layers.

\section{Results} \label{sec:results}

In this section we compare the performance of the proposed multi-level graph convolutional neural network to an analogous full graph GCNN (fGCNN), the original clustered GCNN (oGCNN) with a binary adjacency \cite{vlassis2020geometric}, and for the structured grid data analogous pixel-based CNNs.
The structured grid data also allows comparison with CNN and facilitates filter visualization.

The available data described in \sref{sec:data_sets} is split in a 7:2:1 ratio into training:testing:validation sets.
Using the training and testing data the networks were trained with the Adam stochastic optimizer~\cite{kingma2014adam} with learning rate 0.001 until there were 100 epochs of non-improvement in the mean squared error loss on the test data (up to 1000 epochs).
A batch size of 32 was employed and all input data was normalized to $[0,1]$ prior to training.
All reported (root mean squared) errors and correlations are computed with the held-out validation data.

\subsection{Performance of architecture variants} \label{sec:variants}

\begin{figure}[htb!]
\centering
\begin{subfigure}[b]{0.45\textwidth}
\includegraphics[width=1.0\textwidth]{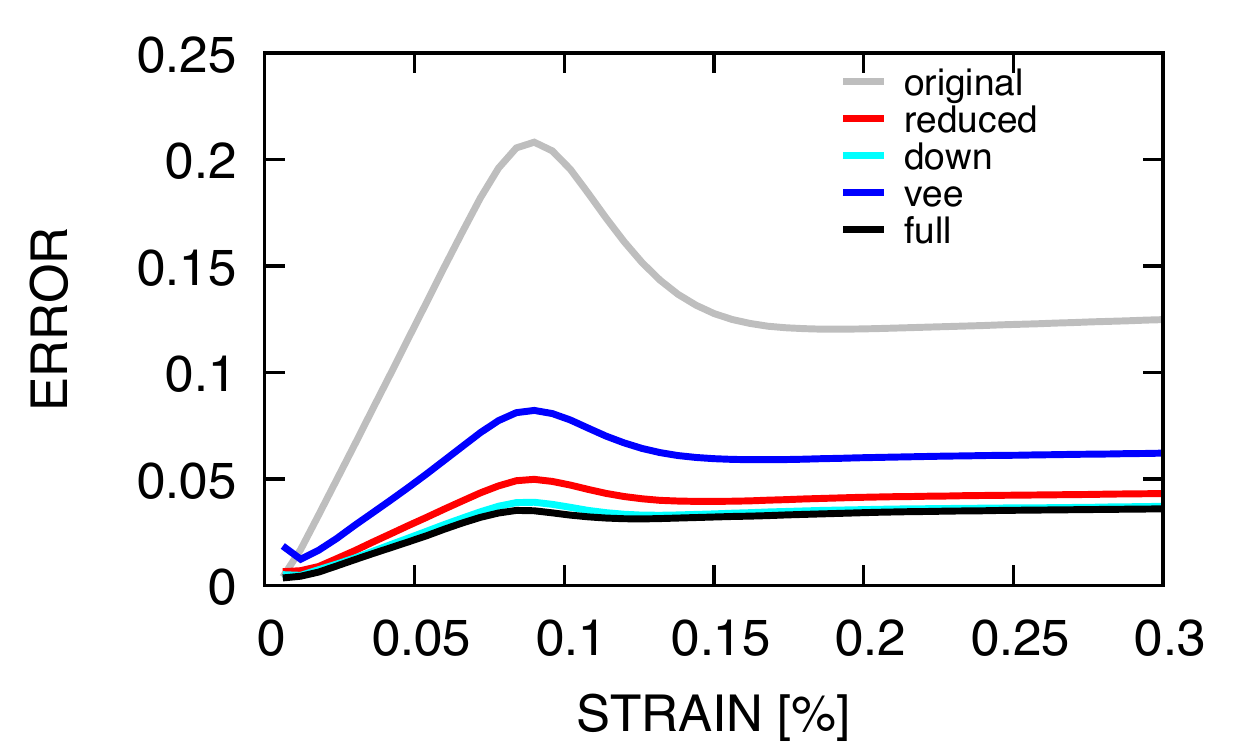}
\caption{Polycrystal error}
\end{subfigure}
\begin{subfigure}[b]{0.45\textwidth}
\includegraphics[width=1.0\textwidth]{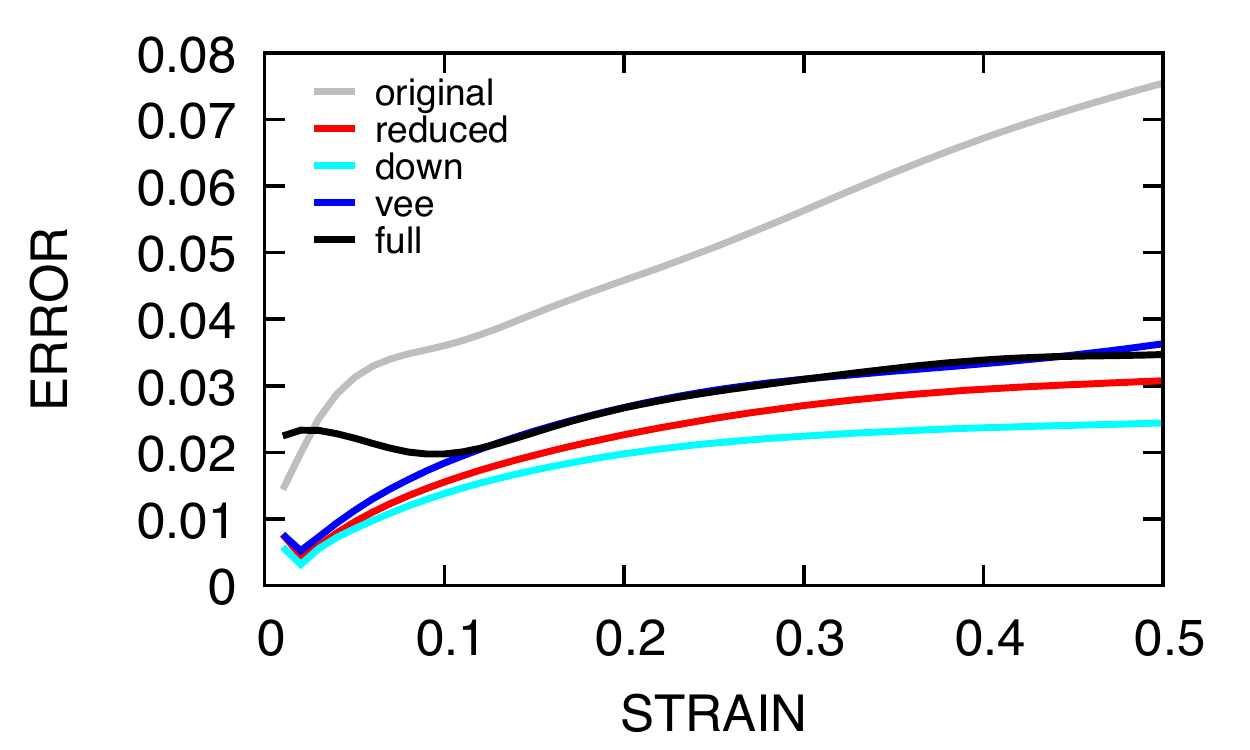}
\caption{Composite error}
\end{subfigure}

\begin{subfigure}[b]{0.45\textwidth}
\includegraphics[width=1.0\textwidth]{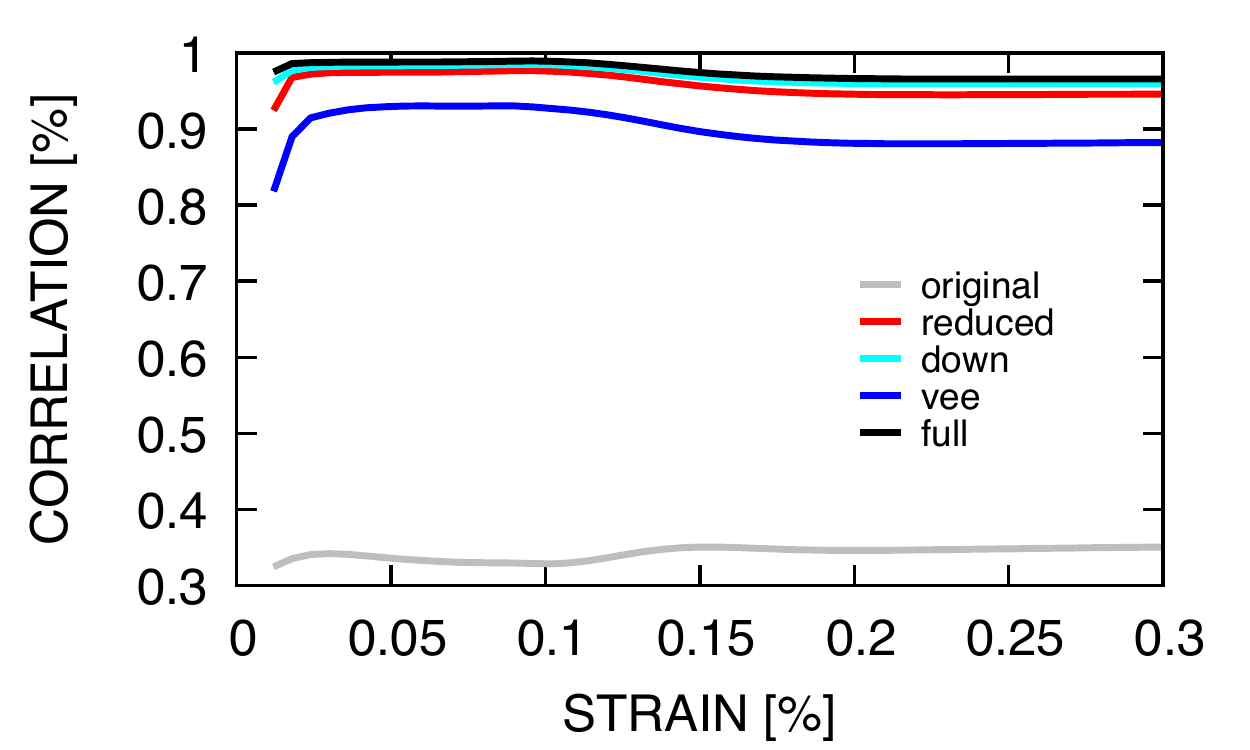}
\caption{Polycrystal correlation}
\end{subfigure}
\begin{subfigure}[b]{0.45\textwidth}
\includegraphics[width=1.0\textwidth]{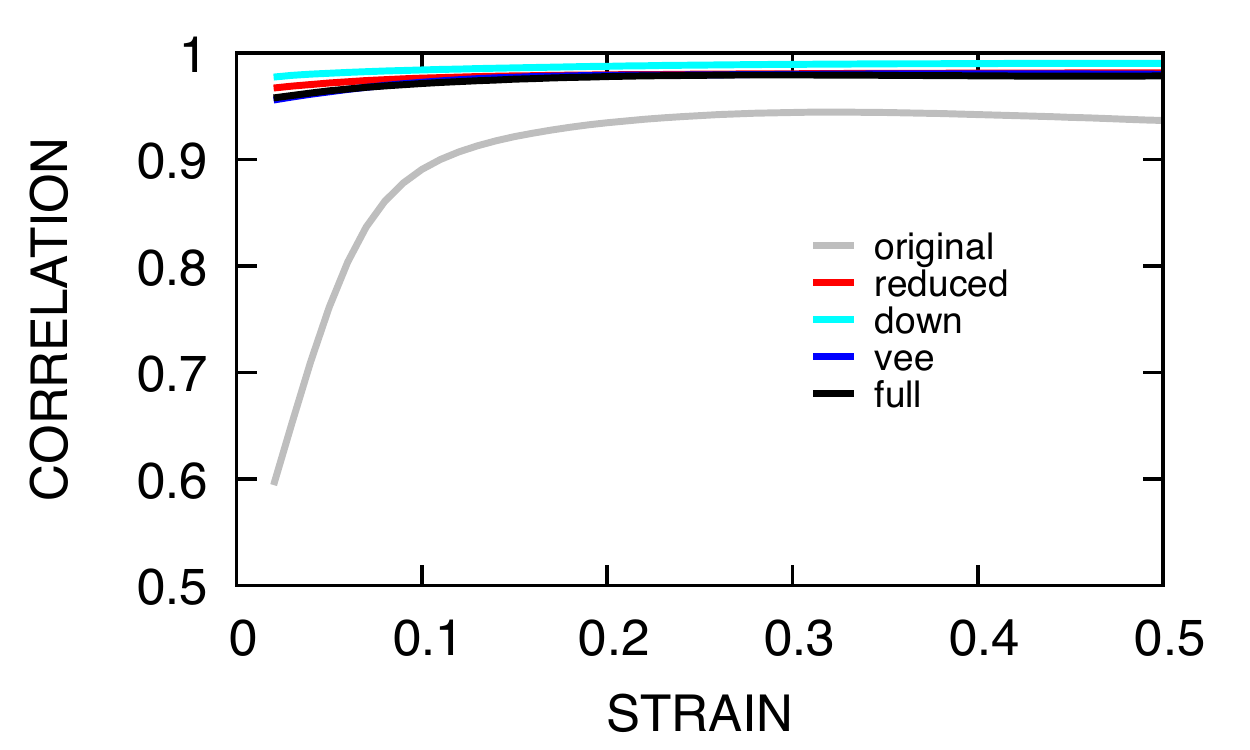}
\caption{Composite correlation}
\end{subfigure}
\caption{Performance of architecture variants on 3D polycrystal plasticity (polycrystal) and viscoelastic matrix with inclusions (composite) datasets.
(a) and (b) show root mean squared error as a function of strain, and (c) and (b) plot the correlation coefficient as a function of strain.
}
\label{fig:variants_performance}
\end{figure}

In \fref{fig:variants_performance} we compare the performance of variants suggested by \fref{fig:general_architecture} with the two large 3D datasets.
The {\it original} architecture refers to a GCNN where the input is only the  clustered data $\Xs^*$ albeit with no attempt at featurization (unlike in \cref{vlassis2020geometric}).
Other four architectures are illustrated in \fref{fig:architecture_variants}.
The {\it full} variant is from \cref{frankel2022mesh} and the other three, namely the vee (vGCNN), down (dGCNN), and reduced (rGCNN), are newly introduced in this paper.
After preliminary, manual tuning the architectures for the polycrystal dataset were given $N_\text{filters} = 16$, $N_\text{conv} =2$ (the total number of convolutional layers) and $N_\text{dense} =3$ (the number of dense layers post global pooling); for the composite dataset hyperparameters were the same, except the number of filters $N_\text{filters} = 8$.
Fully optimizing the hyperparameters was not possible since training these networks took 8-32 hours on a modern GPU.

As shown in \fref{fig:variants_performance}, all variants that incorporate data at the level of the discretization perform better than the original without manual featurization.
The dGCNN variant performs the best among the new variants.
In fact for both exemplar problems the order of performance for the new variants is dGCNN, rGCNN, and vGCNN.
For the polycrystal exemplar the dGCNN and rGCNN variants perform on par with the full variant, while for the composite exemplar both outperform the full variant.
Also apparent is the upscaling of information from cluster to mesh level that is part of the vGCNN does not have advantages for these exemplars.
The fact that the multi-level variants have performance advantages over the full treatment for the composite exemplar (versus the polycrystal exemplar) may be due to the adjacency being one (matrix) to all (inclusions) which requires a more long range interaction to represent inclusion-inclusion effects (versus the contacting crystal-crystal effects).
The down allows for significant mixing of the discretization- and cluster-level information; however, the reduced has the advantage of requiring considerably fewer convolutional operations which for these exemplars the number of elements versus clusters are approximately  10,000:10.

\subsection{Sensitivity to the number of hidden features} \label{sec:hidden}

A primary question regarding deep featurization, particularly with the  {\it reduced} network (rGCNN), is whether the graph reduction prevents the discovery of all the features necessary to predict the output of interest.
Specifically by choosing $N_\text{filters}$ and $N_\text{features}$, refer to \fref{fig:architecture}, large enough can we create a rGCNN with performance on par with a full GCNN.
This study shows the convergence with the number of filters and serves to set the architecture for the subsequent study.
We perform two investigations, one for the 2D unstructured heat flux data and another for the 2D structured crystal plasticity data.

\begin{figure}[htb!]
\centering
\begin{subfigure}[b]{0.45\textwidth}
\includegraphics[width=1.0\textwidth]{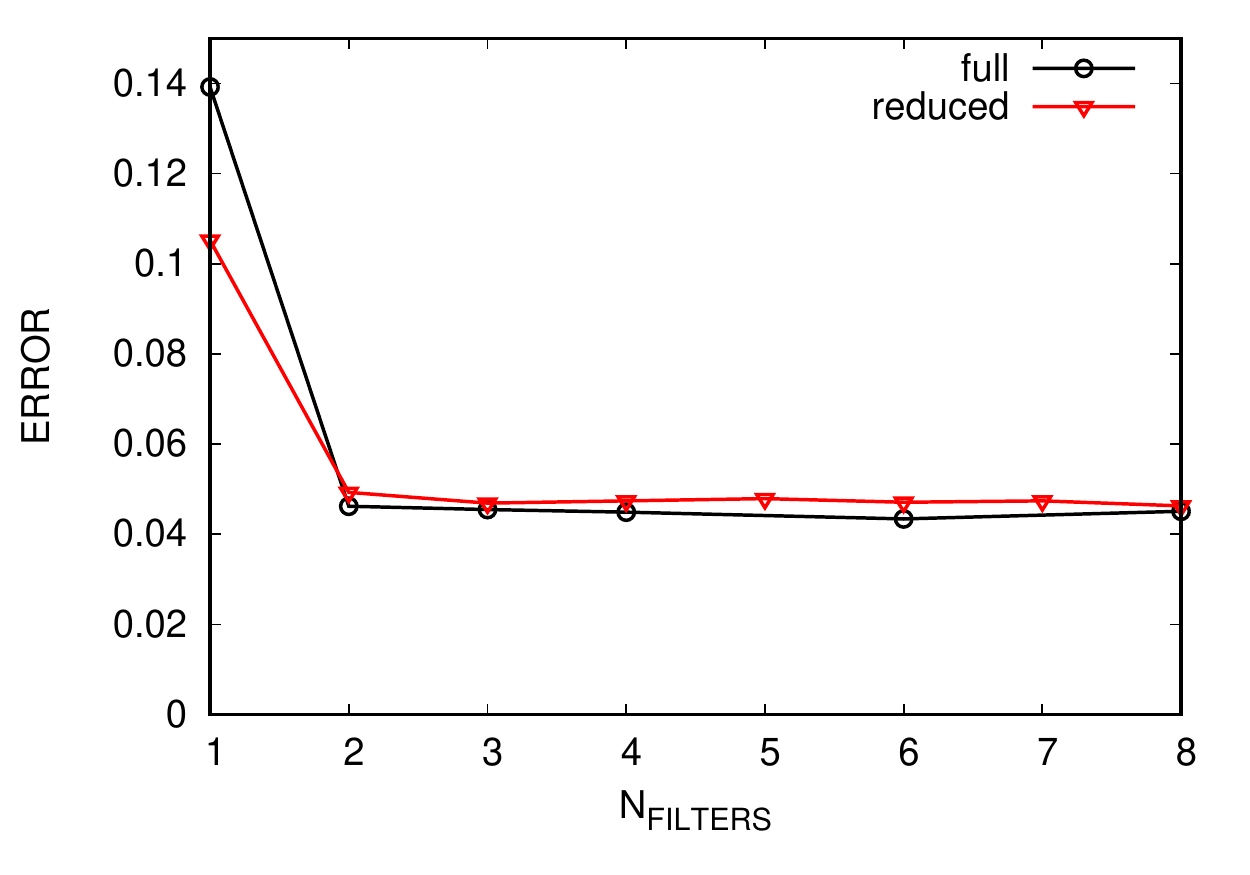}
\caption{Heat flux}
\end{subfigure}
\begin{subfigure}[b]{0.45\textwidth}
\includegraphics[width=1.0\textwidth]{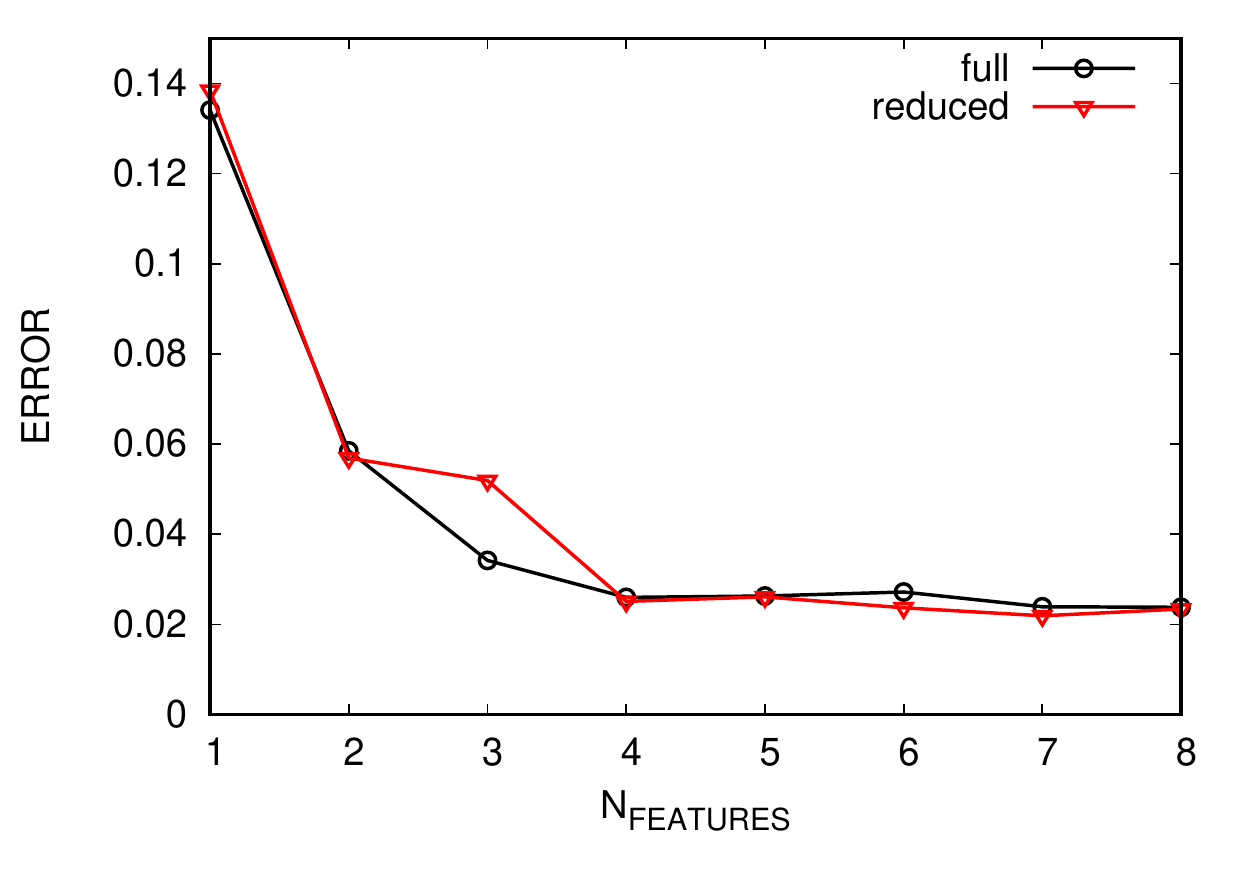}
\caption{Stress}
\end{subfigure}
\caption{
Sensitivity to number of filters and features.
Heat flux (a)
for both the fGCNN and rGCNN $N_\text{filters}$ is varied, andfor the rGCNN $N_\text{features}=3$ fixed.
Stress (b): $N_\text{features} = N_\text{filters}$ is varied for both.
}
\label{fig:nfeatures}
\end{figure}

For the heat flux data study we first vary the number of filters that determine $N_\text{features}$ in the full GCNN.
\fref{fig:nfeatures}a shows that the performance improves up to 2-3 features and then it remains more or less flat.
Since this indicates 3 features are sufficient, for the rGCNN $N_\text{features}$ (the number of filters in the post-featurization convolutional stack) is fixed at 3 and $N_\text{filters}$ which determines the embedding is varied.
For this scenario, \fref{fig:nfeatures}a shows that $N_\text{filters}=N_\text{features}$ is sufficient to produce equivalent performance to the fGCNN.
For this data the size of the output MLP is  $N_\text{dense} = 2$, including the linear mixing layer.

For the crystal plasticity study, \fref{fig:nfeatures}b shows the convergence of the rGCNN for $N_\text{filters} = N_\text{features}$ as well as the convergence for the corresponding fGCNN.
A decrease in improvement with increasing complexity for each configuration is achieved with $N_\text{features} \ge 4$.
It is not surprising that more hidden features are necessary to represent the more complex process.
For this data only a linear mixing layer ($N_\text{dense}=1$) processes the global features before they are sent to the long short-term memory (LSTM) \cite{hochreiter1997long} recurrent neural network (RNN) unit.

The fact that performance of these networks is tolerant to redundancy will be revisited in \sref{sec:viz}.

\begin{figure}[htb!]
\centering
\begin{subfigure}[b]{0.45\textwidth}
\includegraphics[width=1.0\textwidth]{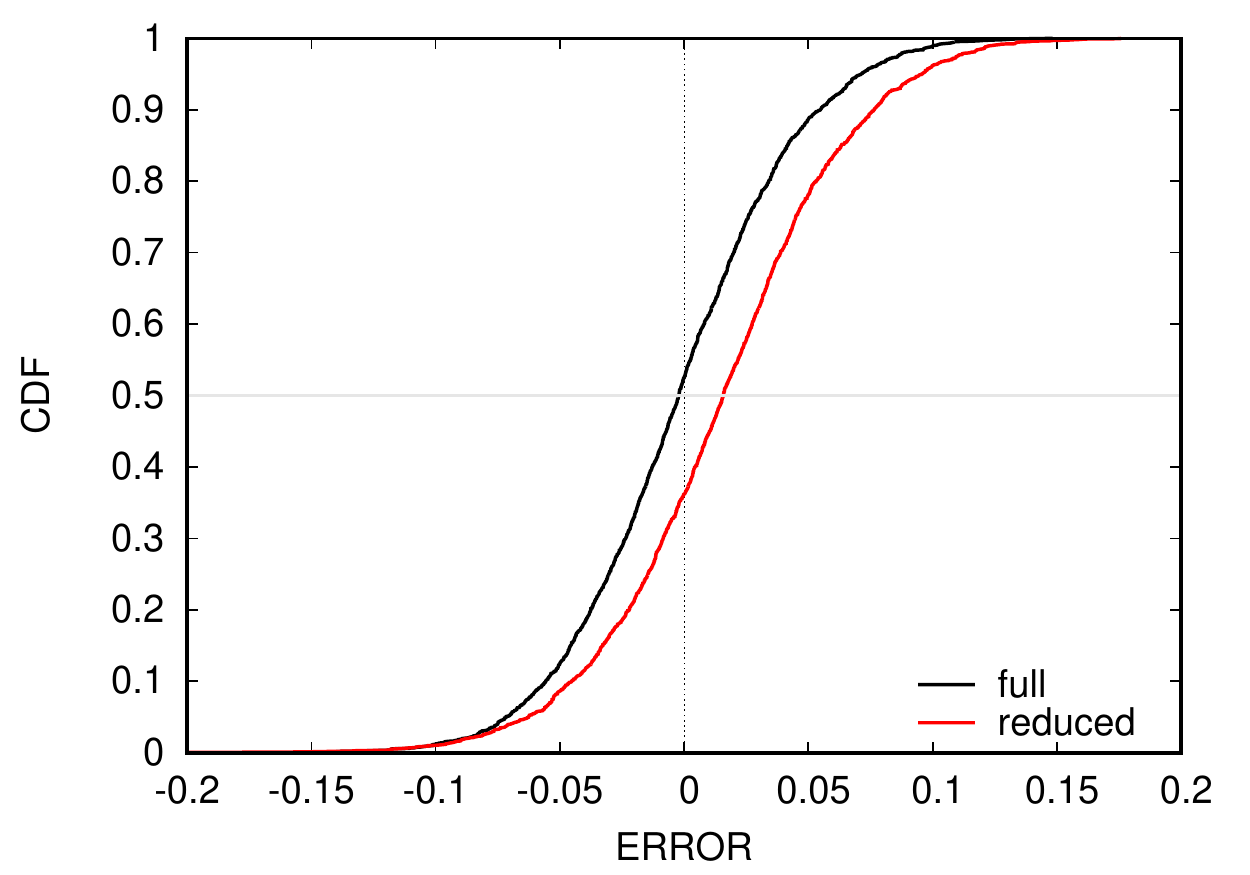}
\caption{Heat flux}
\end{subfigure}
\begin{subfigure}[b]{0.45\textwidth}
\includegraphics[width=1.0\textwidth]{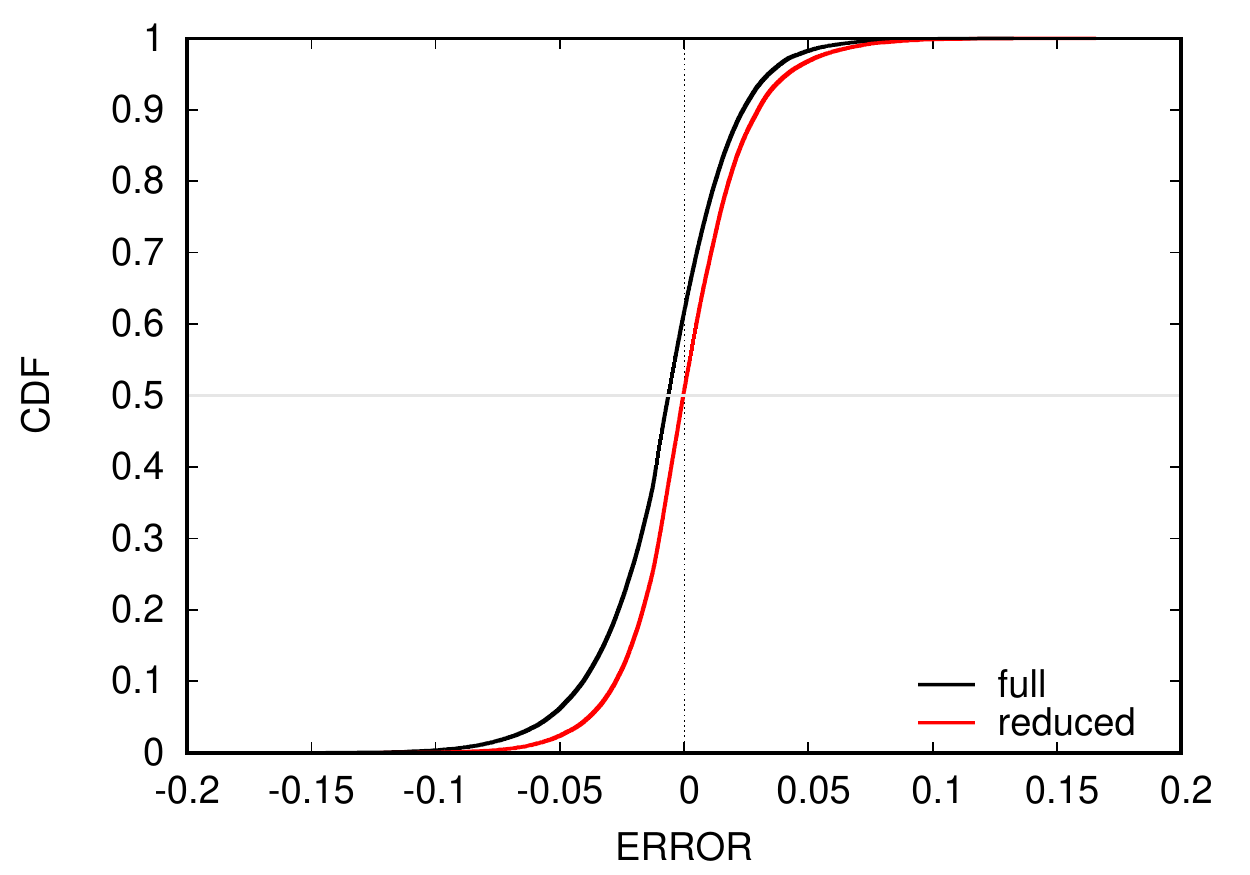}
\caption{Stress}
\end{subfigure}
\caption{fGCNN and rGCNN performance comparison for heat flux and plastic stress exemplars.
}
\label{fig:cdf_comp}
\end{figure}

For the optimal architectures ($N_\text{features} = 3$ and $N_\text{filters} =2$ for the heat flux example and $N_\text{features} = 3$ and $N_\text{filters} =3$ for the stress example) \fref{fig:cdf_comp} shows the distribution of errors for: (a) conductivity data on an unstructured mesh and (b) the stress evolution predictions using a structured mesh.
The errors for the rGCNN are on par with the fGCNN for the heat flux example and slightly better for the more complex stress data.
Also the rGCNN exhibits less bias than the fGCNN on the more complex data.

\subsection{Interpretation} \label{sec:viz}

Using the correlation and visualization techniques described in \sref{sec:interpret} we can provide some insight into how the GCNNs are representing the response functions.
Here we focus only on the CP response since we also use the CNN architecture for comparison and the response goes through different regimes (i.e. elastic-yield-plastic flow).
Note the CNN has much richer kernels (9 independent parameters for a kernel width of 3,  instead of 2 for the GCNN with self and first neighbor weights).

\begin{figure}[htb!]
\centering

\begin{subfigure}[b]{0.55\textwidth}
\includegraphics[width=0.95\textwidth]{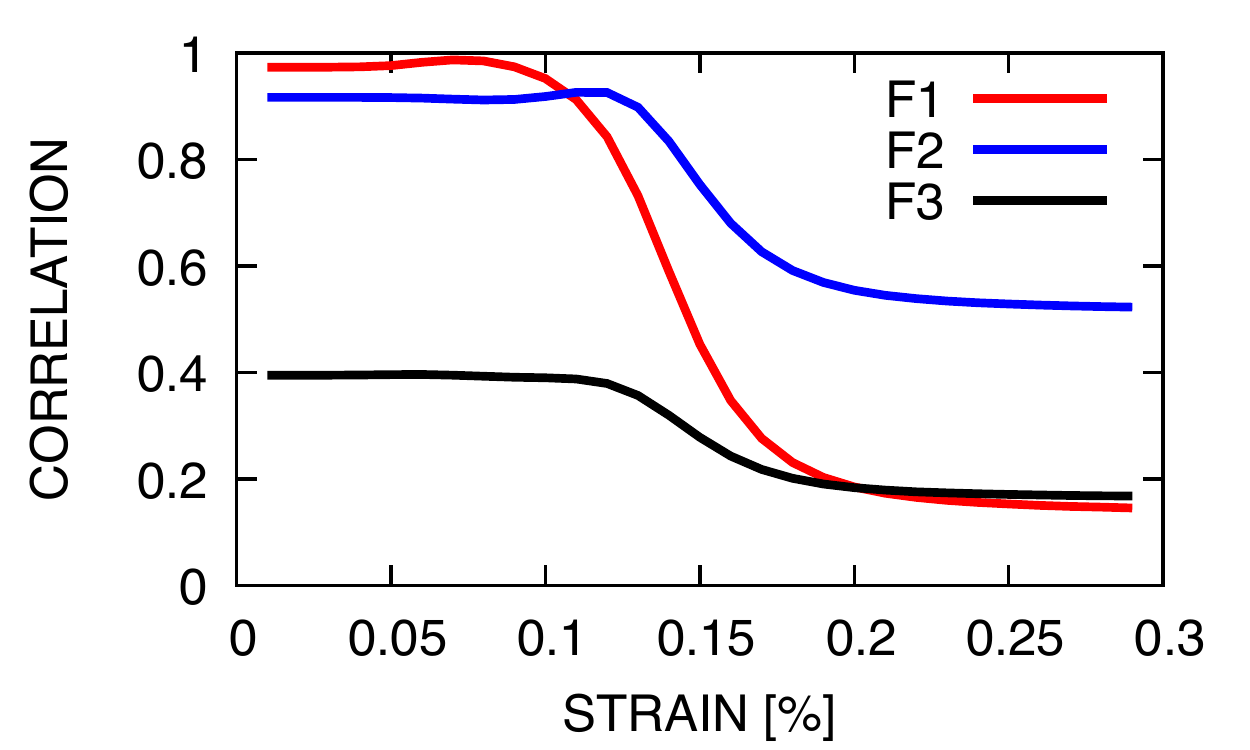}
\caption{CNN}
\end{subfigure}

\begin{subfigure}[b]{0.55\textwidth}
\includegraphics[width=0.95\textwidth]{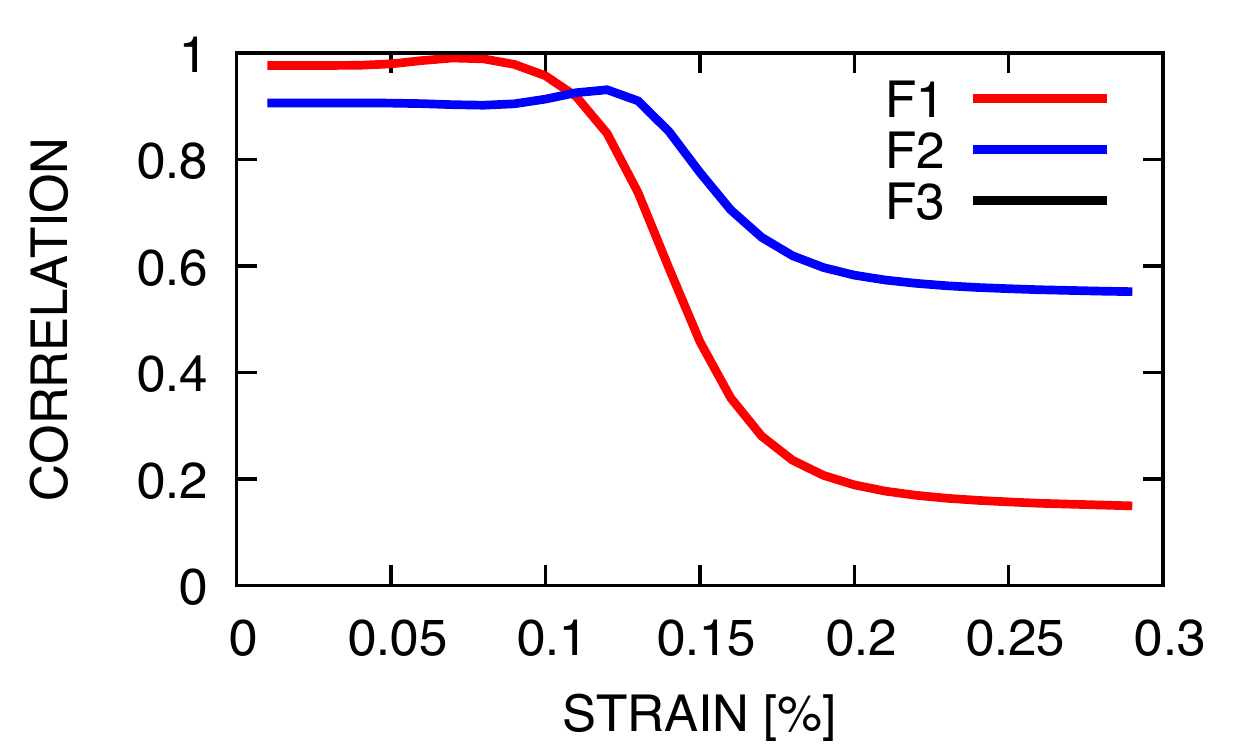}
\caption{GCNN}
\end{subfigure}

\begin{subfigure}[b]{0.55\textwidth}
\includegraphics[width=0.95\textwidth]{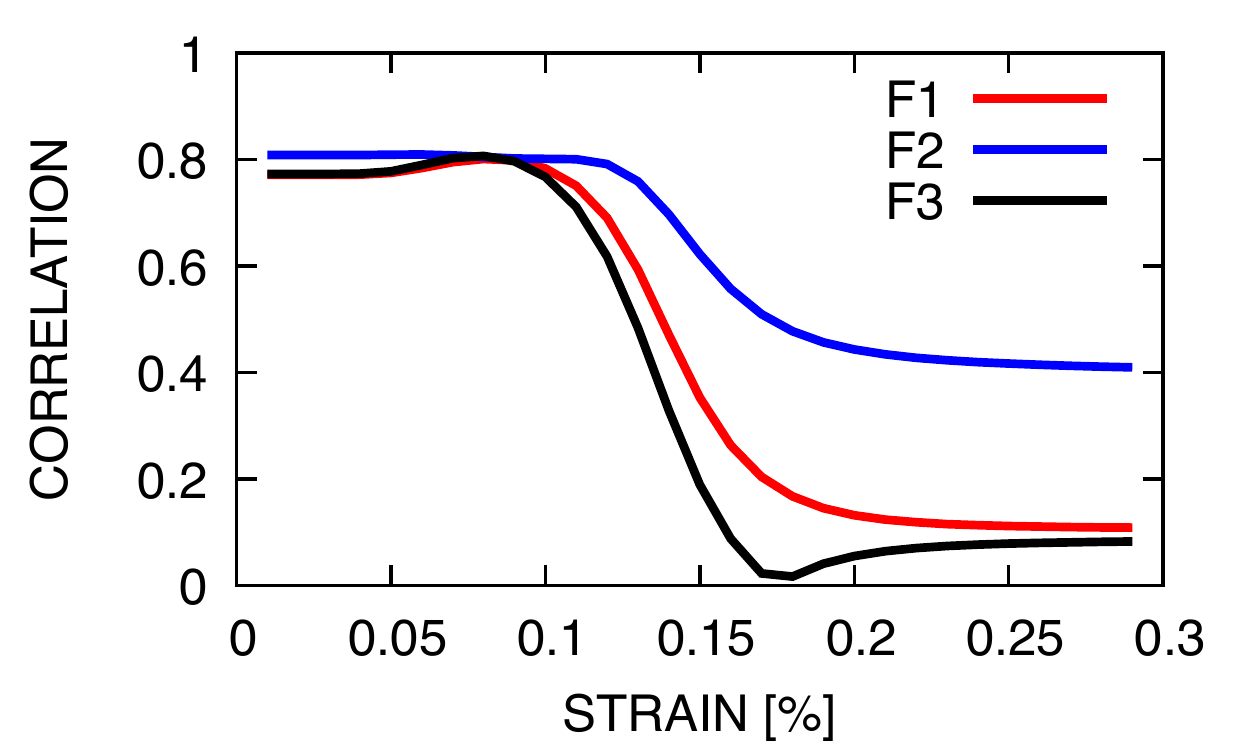}
\caption{rGCNN}
\end{subfigure}

\caption{Global pooling feature $F_a$ correlation with output stress $\bar{\sigma}(t)$.
}
\label{fig:feature_correlation}
\end{figure}

\fref{fig:feature_correlation} shows the correlation of the output of the three filters $F_a$ with the ultimate output of the network $\bar{\sigma}(t)$ as a function of applied strain $\bar{\epsilon}(t)$.
Clearly certain filters correlate with the output more strongly in certain regimes and are likely weighted more heavily in those regimes by the post-GCNN/pre-RNN mixing layer, refer to \fref{fig:architecture}.
The transition in importance occurs at the elastic-plastic transition for the ensemble, refer to \fref{fig:cp_correlations}a, with the low strain regime being elastic and the higher strain associated with plastic flow.
Remarkably the GCNN replicates the correlations of two of the three CNN filters precisely, the third GCNN filter is not active which is likely due to the richer kernel space of the CNN not being representable in a GCNN filter \cite{frankel2022mesh}.
The rGCNN has similar trends albeit different levels of correlation throughout the stress evolution.
It appears to use its filters in a more coordinated way in order to represent the output of interest.

Regarding filter activation for the ensemble of data all filters in the CNN convolutional stack produce non-zero output, whereas for the GCNN in both the first and the second layers only two of the three are active.
This is somewhat surprising but it appears that the training is producing sparsity induced by redundancy in the information the filters can produce.
For the rGCNN all filters in the stack are active but appear to be producing redundant output.

\begin{figure}[htb!]
\centering

\begin{subfigure}[b]{0.95\textwidth}
\includegraphics[width=0.30\textwidth]{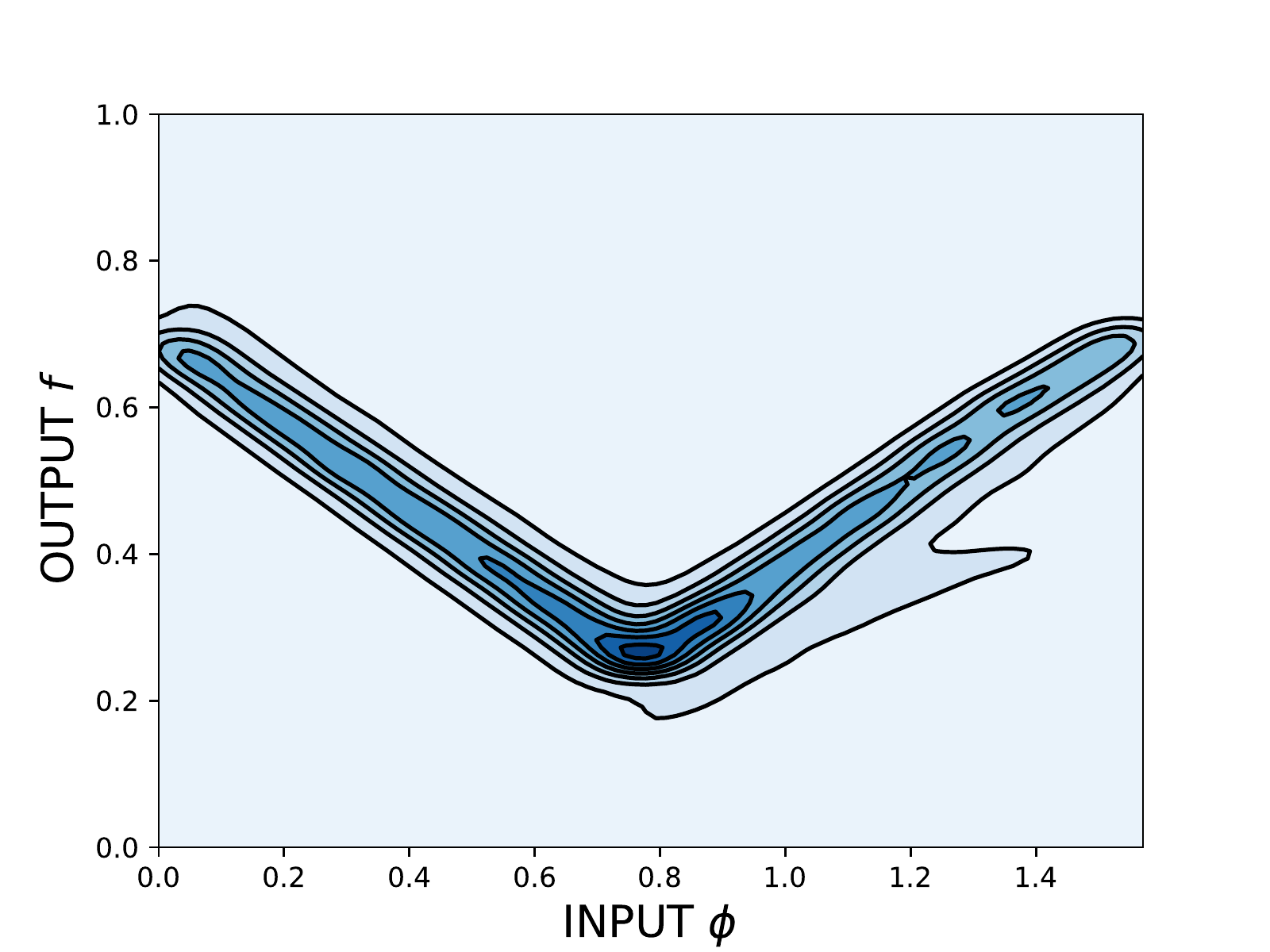}
\includegraphics[width=0.30\textwidth]{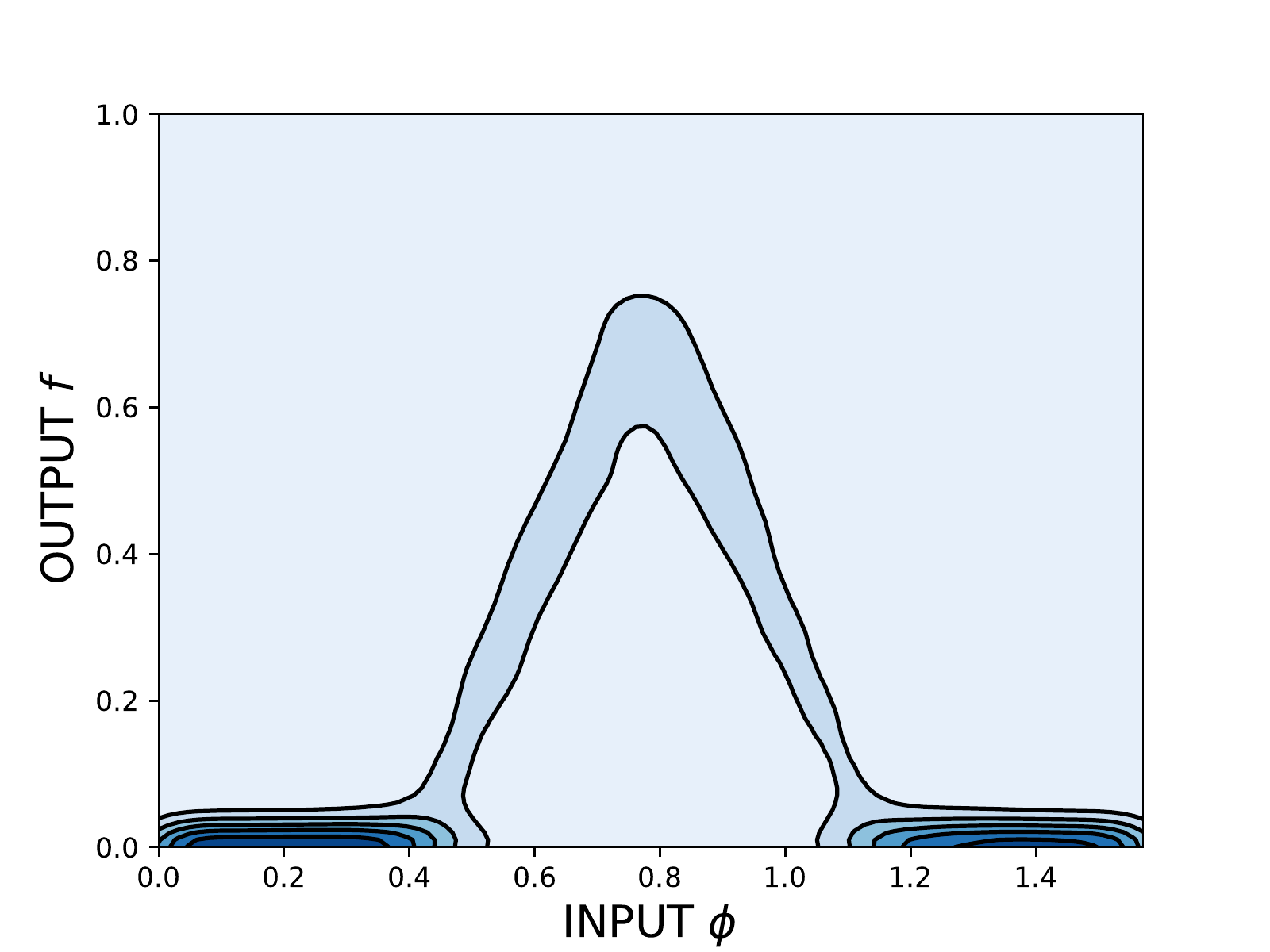}
\includegraphics[width=0.30\textwidth]{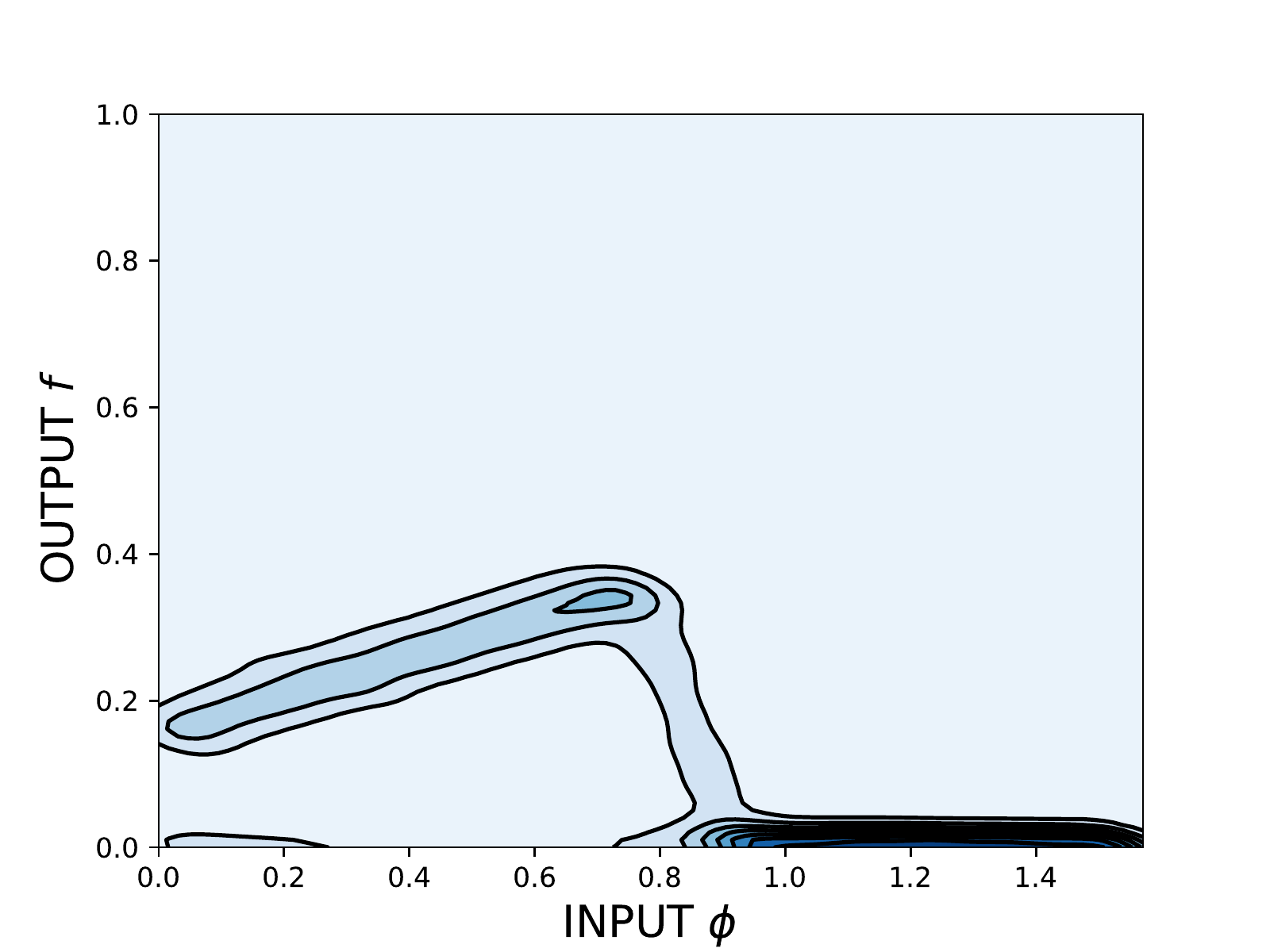}
\caption{CNN}
\end{subfigure}

\begin{subfigure}[b]{0.95\textwidth}
\includegraphics[width=0.30\textwidth]{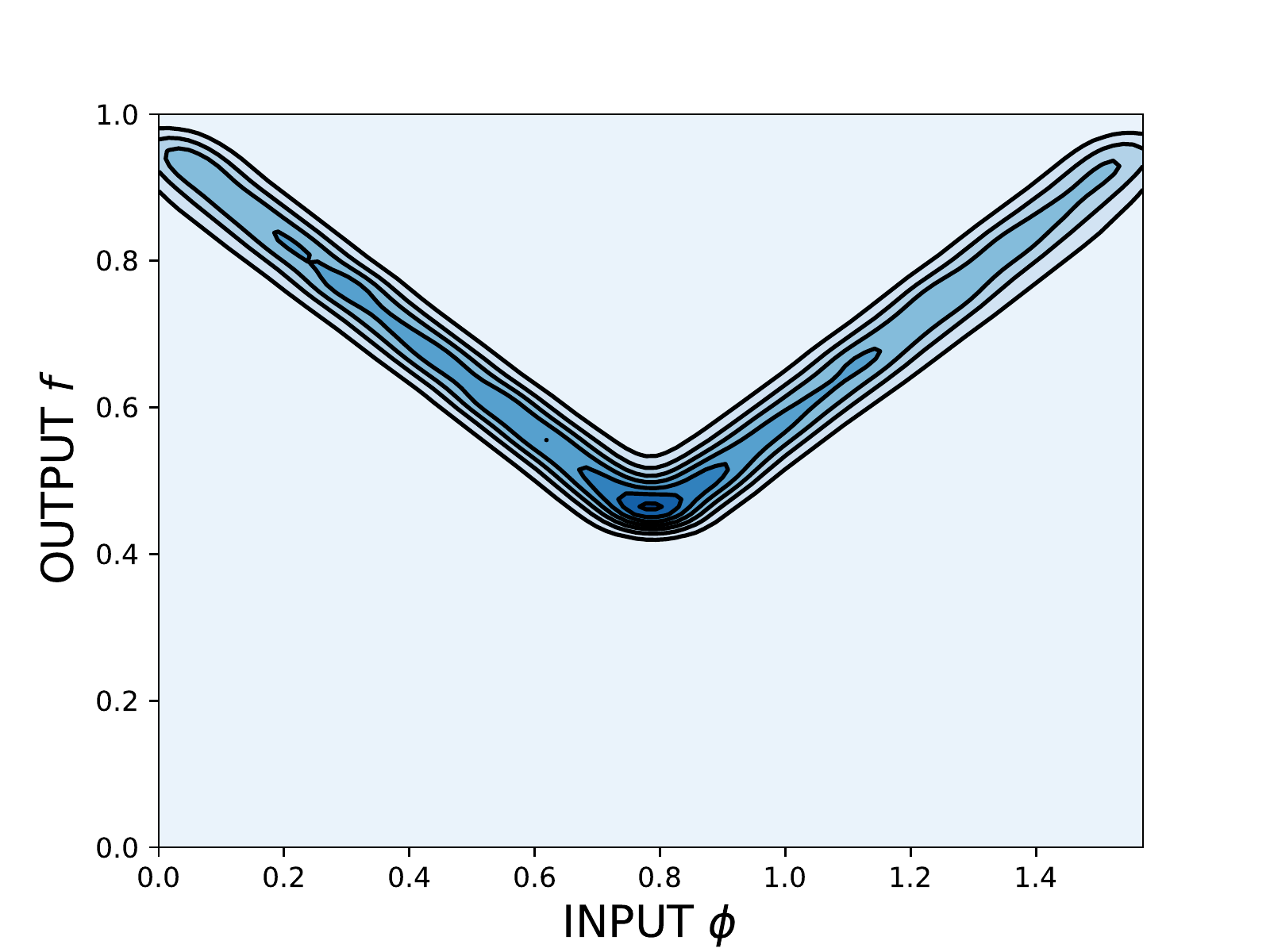}
\includegraphics[width=0.30\textwidth]{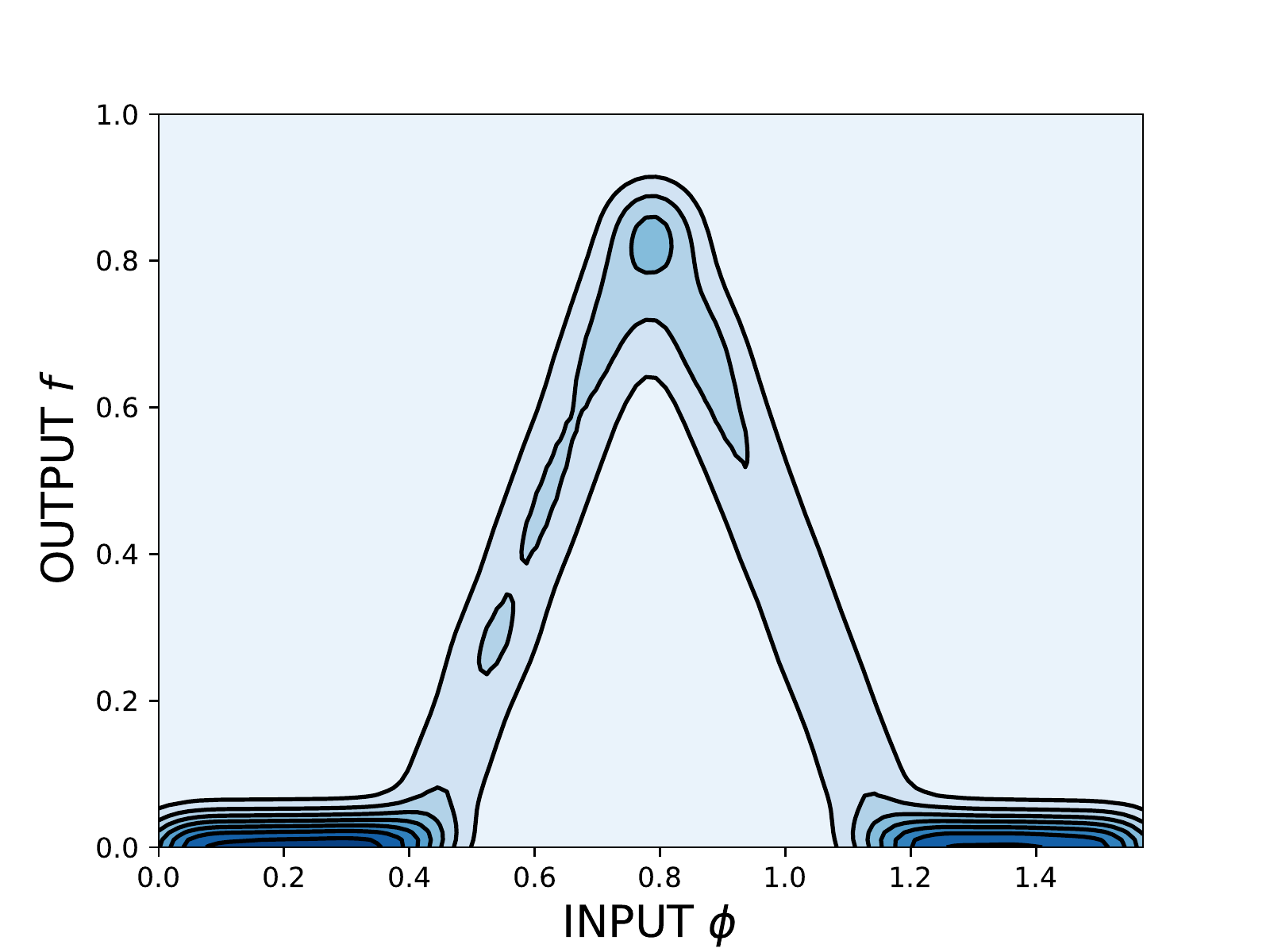}
\includegraphics[width=0.30\textwidth]{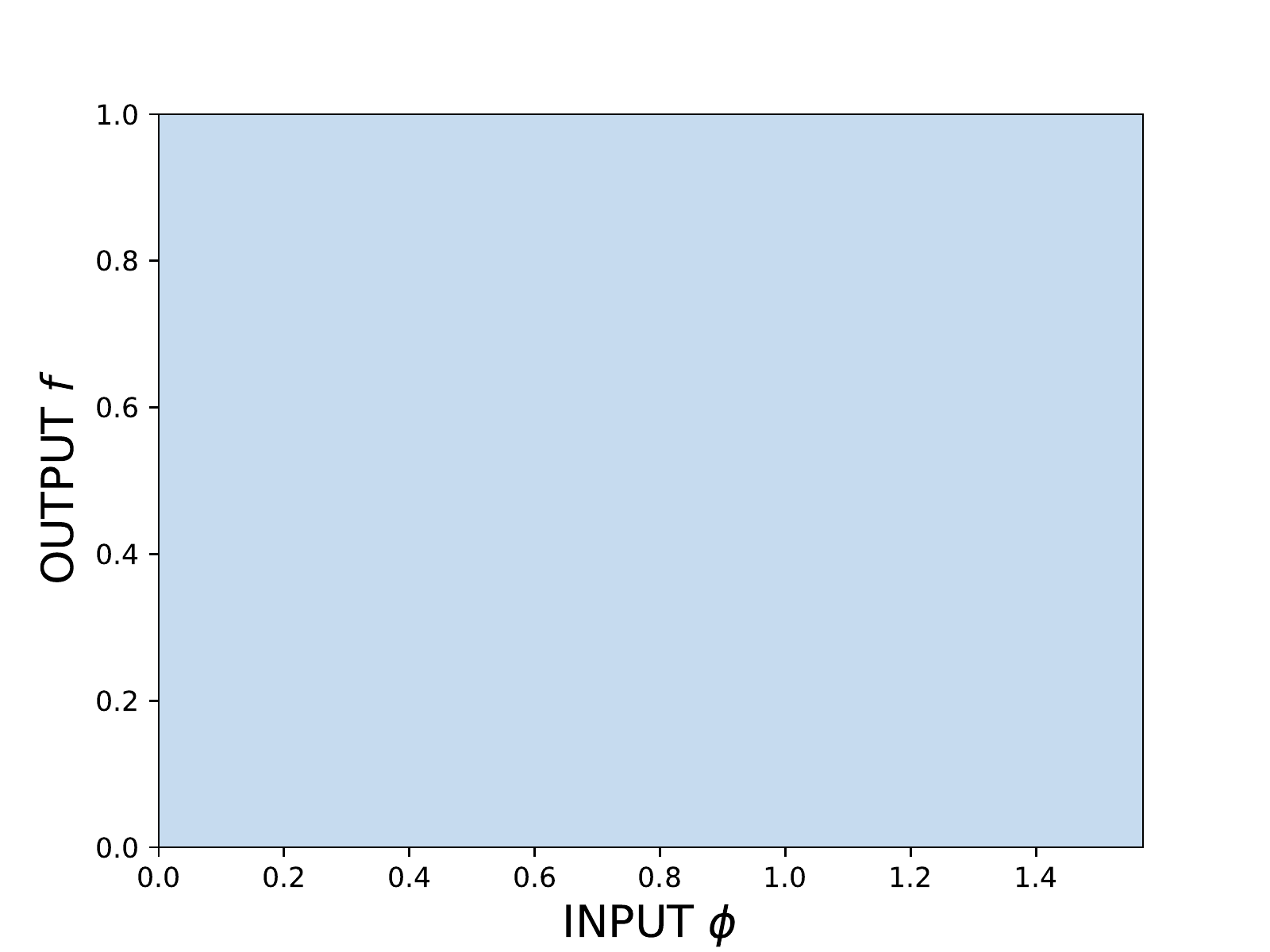}
\caption{GCNN}
\end{subfigure}

\begin{subfigure}[b]{0.95\textwidth}
\includegraphics[width=0.30\textwidth]{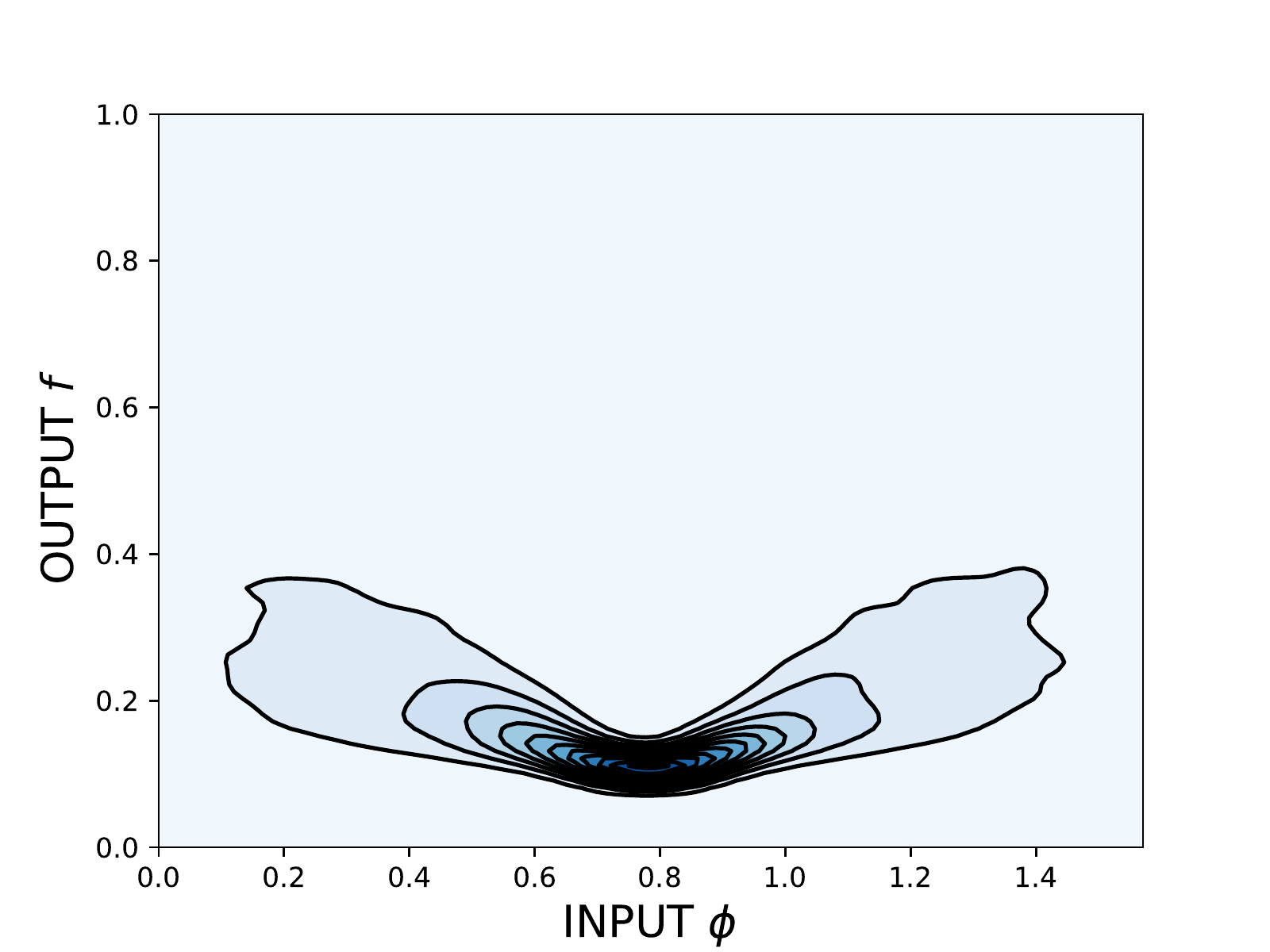}
\includegraphics[width=0.30\textwidth]{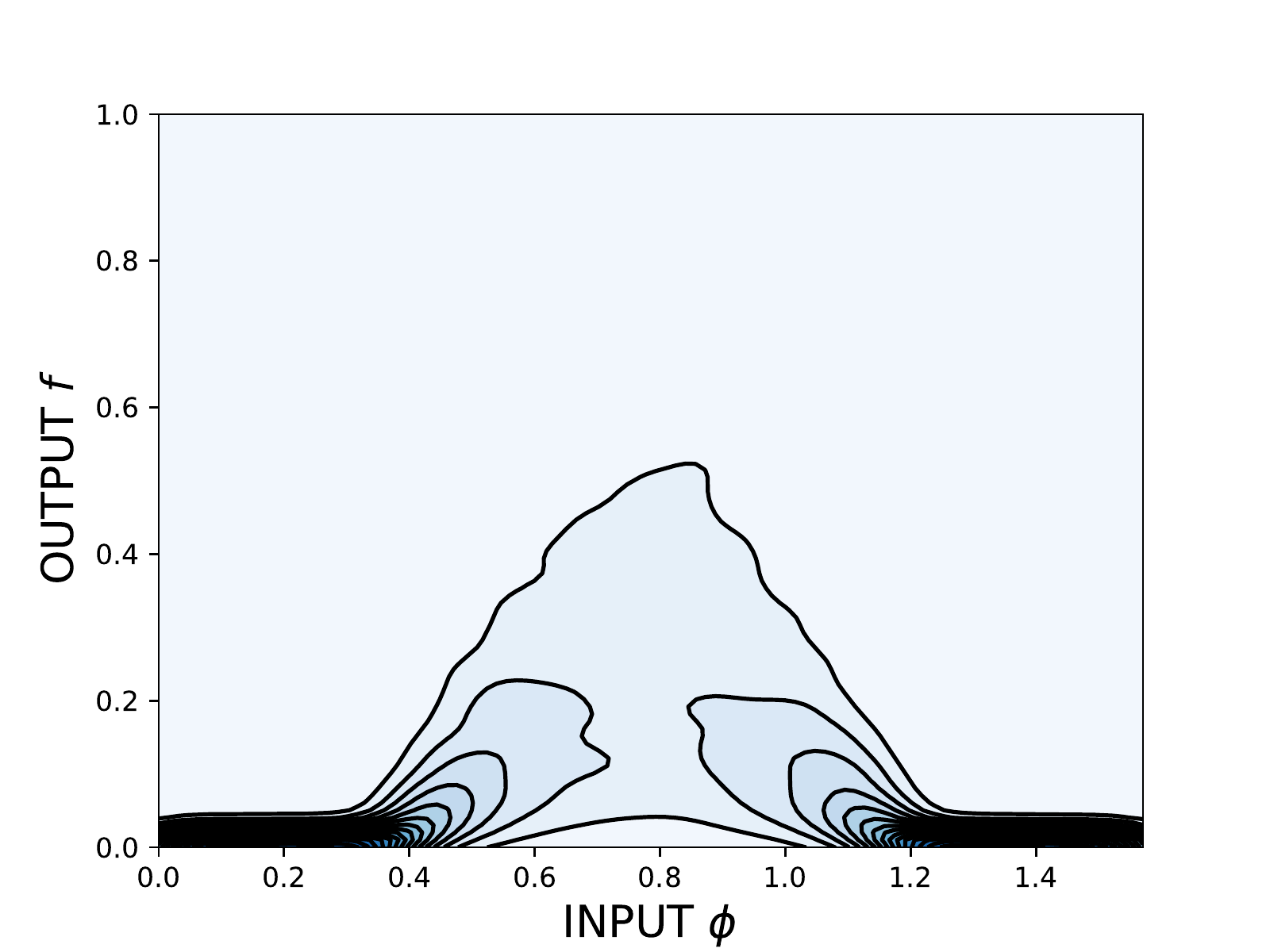}
\includegraphics[width=0.30\textwidth]{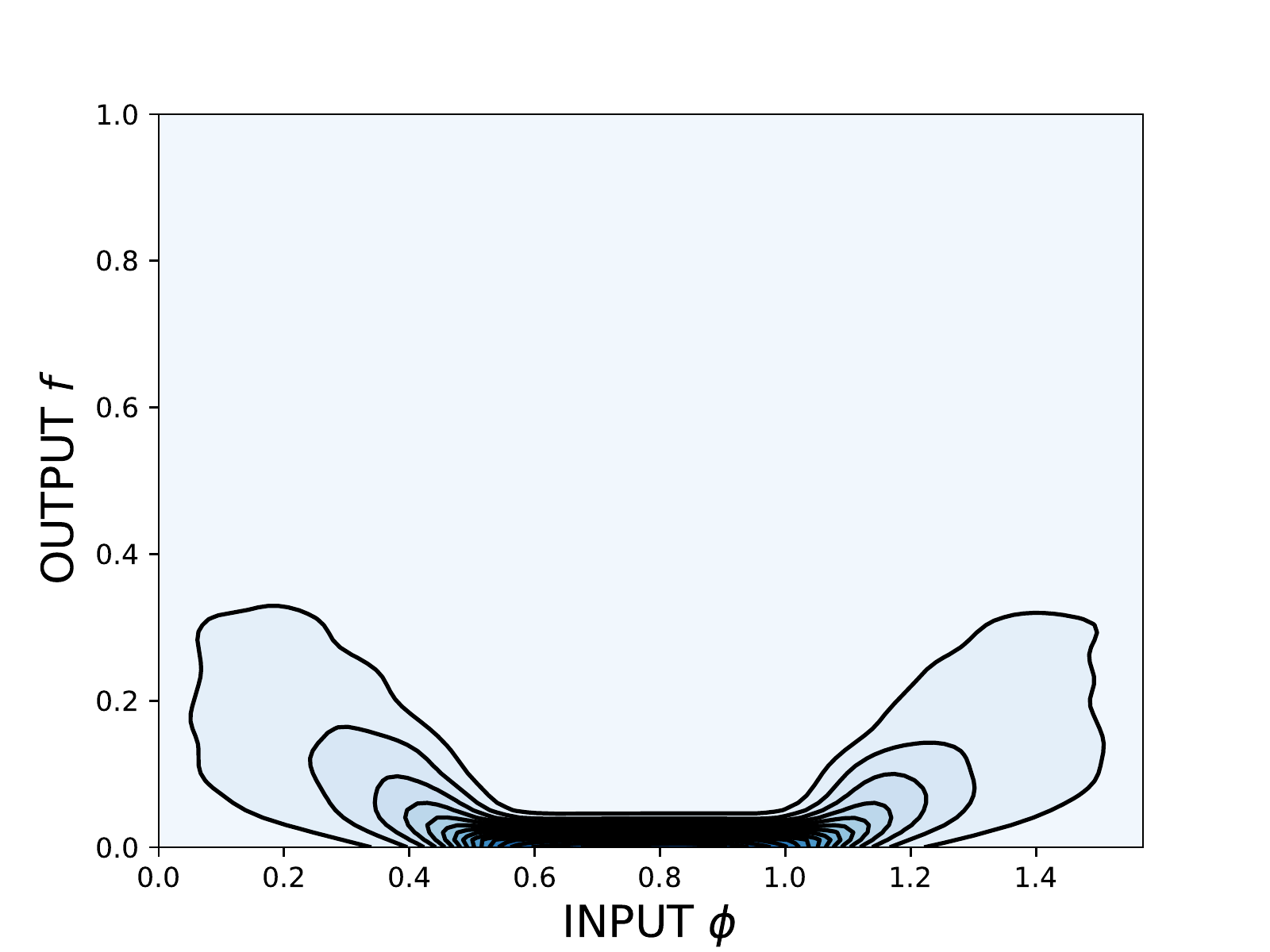}
\caption{rGCNN}
\end{subfigure}

\caption{Filter maps $\phi_i \to Z_{ia}$, for each pixel $i$ and filter $a$, shown as density plots.
}
\label{fig:filter_maps}
\end{figure}

The density plots in \fref{fig:filter_maps} show the correlation of input $\phi_i$ and filter output $Z_{ia}$ per node (or pixel).
Here $i$ indexes nodes and $a$ indexes filters.
Clearly there is some similarity in the features $Z_{ia}$ each network are producing.
The primary filter (left column in \fref{fig:filter_maps}) in the elastic regime folds most of the grain orientations $\phi$ through a periodic bilinear function, similar to $\cos(2\phi)$.
This is consistent with expectations from analytic homogenization models, refer to \sref{sec:problem}.
The primary plastic filter (middle column in \fref{fig:filter_maps}) is more of a tent function on input values, therefore focuses on a sub-range of inputs.
It appears that this filter is placing the most importance on the orientations most aligned with loading.
The third filter produces qualitatively different output for each architecture: for the CNN it only produces non-zero output for a different range, for the GCNN it is off, and for the rGCNN it appears to produce output loosely complementary to the second filter.
It is also interesting that the scatter in the CNN and GCNN maps is remarkably smaller than for the rGCNN, which seems to be a signature of the relative coordination between the filters in producing the output of interest.

\begin{figure}[htb!]
\centering
\begin{subfigure}[b]{0.45\textwidth}
\includegraphics[width=0.95\textwidth]{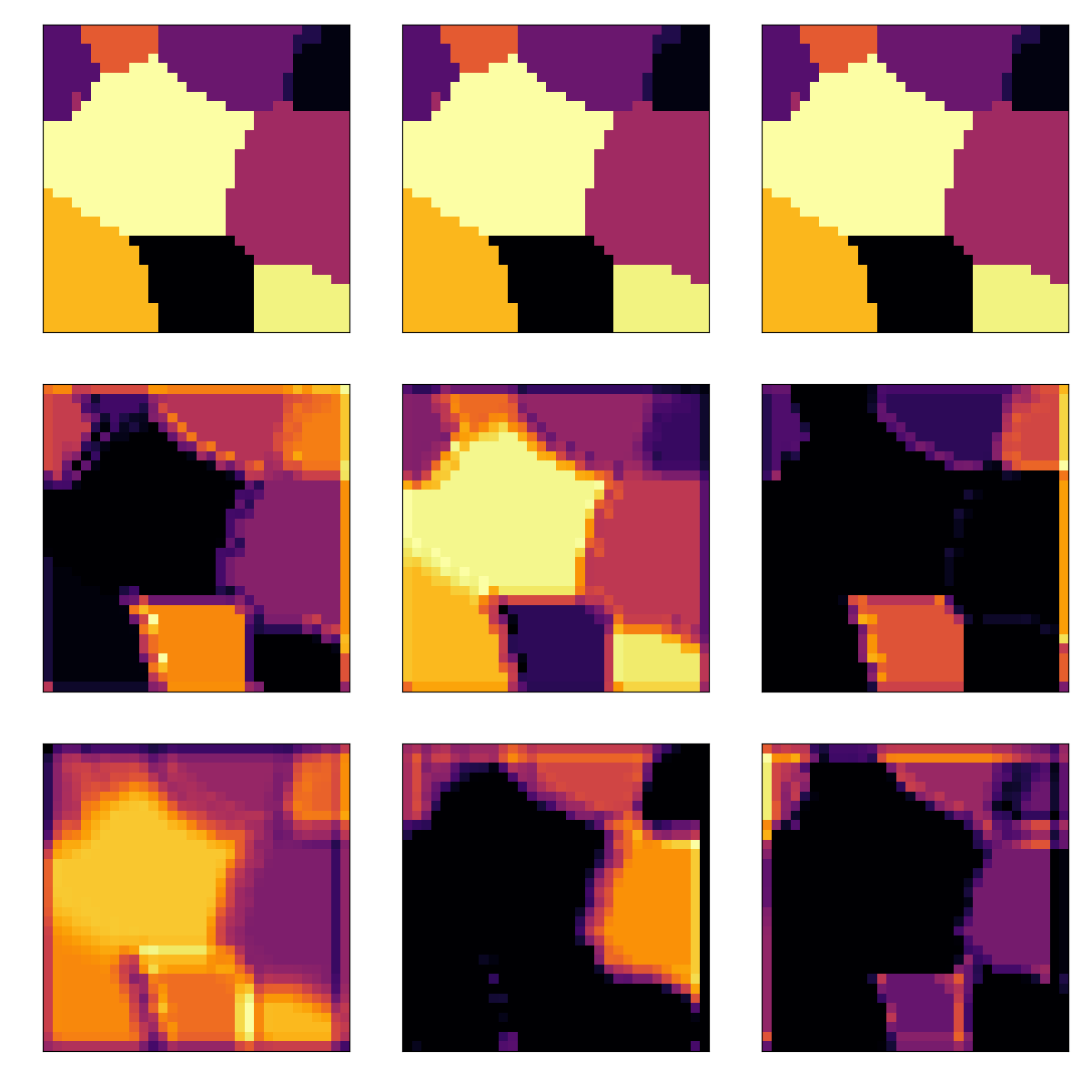}
\caption{CNN}
\end{subfigure}
\begin{subfigure}[b]{0.45\textwidth}
\includegraphics[width=0.95\textwidth]{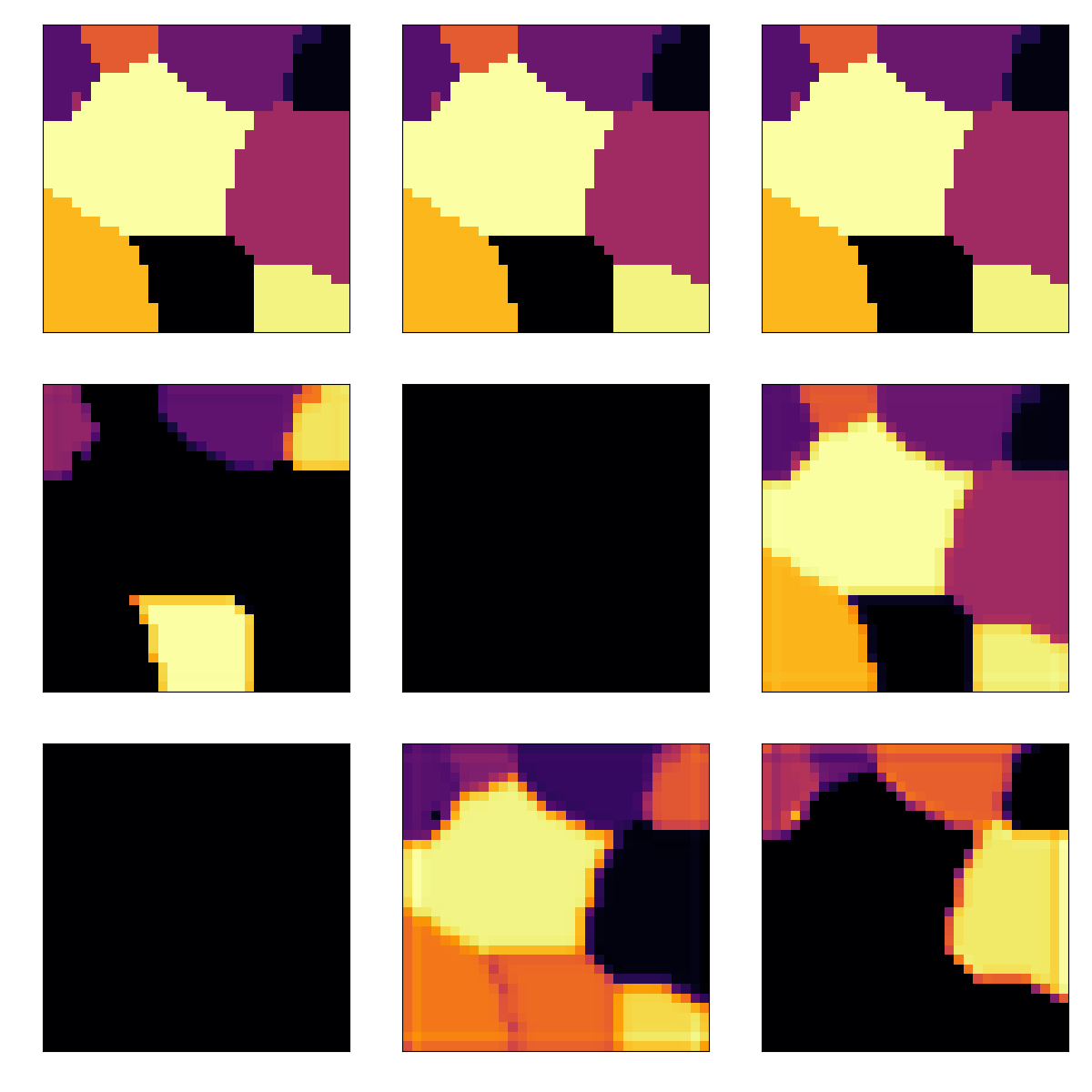}
\caption{GCNN}
\end{subfigure}
\caption{Filter visualization for a representative sample (top row: image of microstructure) through the filter stack (lower rows) for the (a) CNN and (b) GCNN architectures.
}
\label{fig:gcnn_filter_viz}
\end{figure}

\begin{figure}[htb!]
\centering
\begin{subfigure}[b]{0.45\textwidth}
\includegraphics[width=0.95\textwidth]{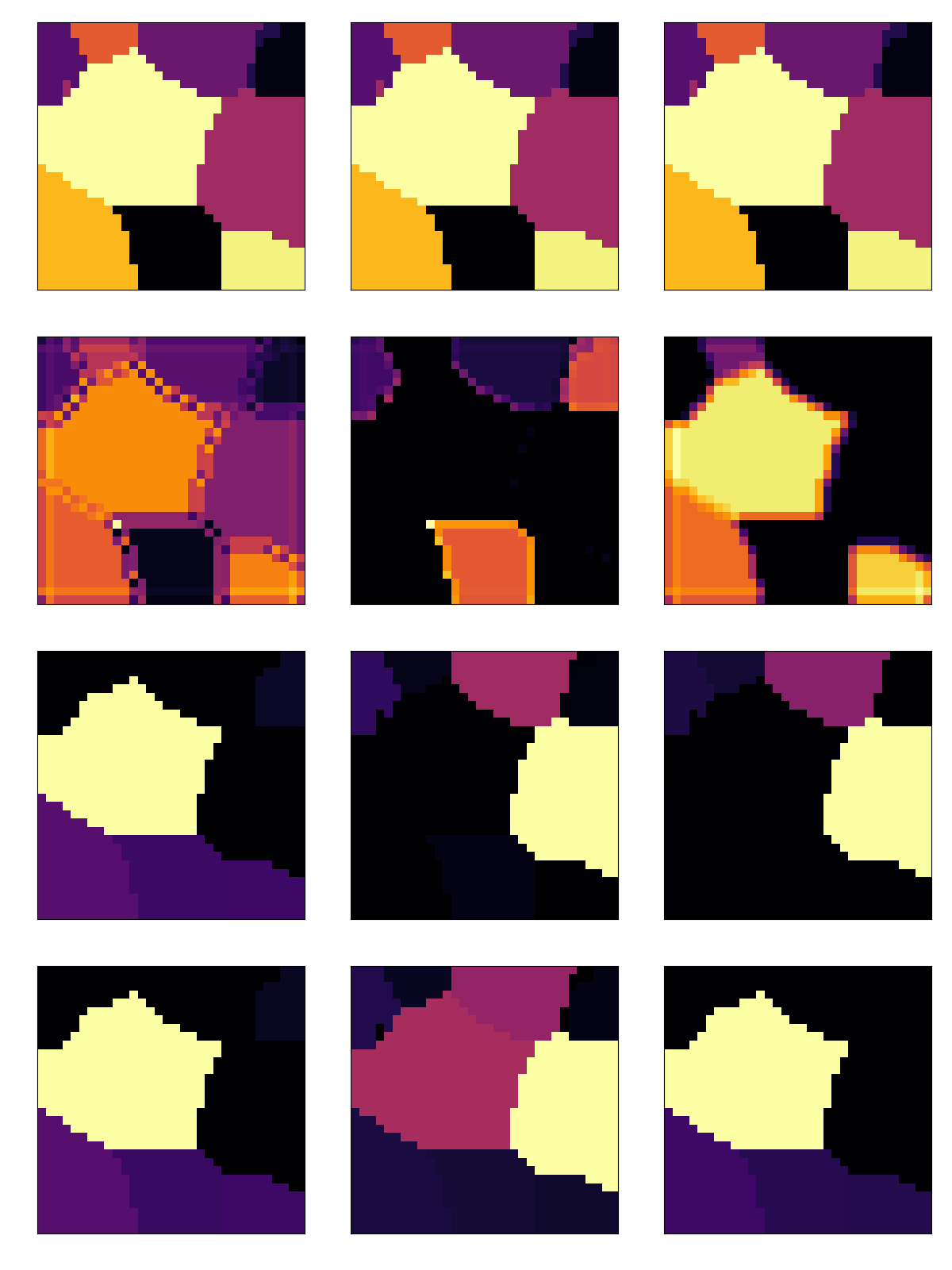}
\caption{rGCNN, RMSE=0.0247}
\end{subfigure}
\begin{subfigure}[b]{0.45\textwidth}
\includegraphics[width=0.95\textwidth]{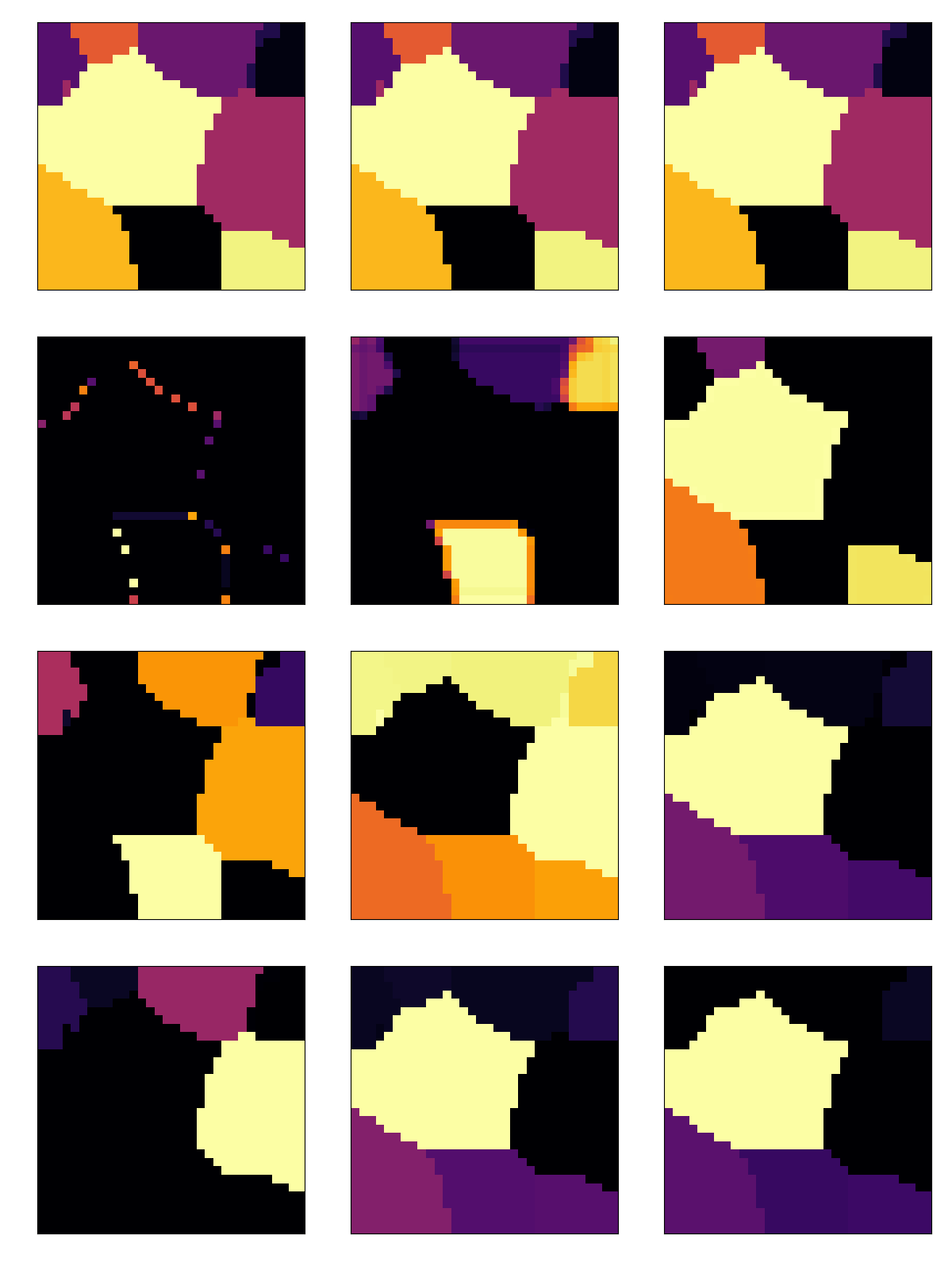}
\caption{rGCNN, RMSE=0.0340}
\end{subfigure}
\caption{Filter visualization for a representative sample (top row) through the filter stack (lower rows) for (a) rGCNN with RMSE=0.0247 and (b) rGCNN with RMSE=0.0340.
}
\label{fig:rgcnn_filter_viz}
\end{figure}

Lastly, \fref{fig:gcnn_filter_viz} shows the output of each filter in the convolutional layer stack given the input on the top row.
Some regions produce low values (black) while other activate high values (yellow).
There is an approximate correspondence between the filter outputs of the two architectures.
As discussed folding of angles to reflect the symmetry of the output and/or emphasizing of orientations most aligned with the loading create these different activations.
What is not apparent in the previous analysis is there are some signatures of local edge detection, particularly in the last outputs which are effective 2-hops from the input (the middle row effects a 1-hop diffusion).
Notably it appears the GCNN places less emphasis on the inter-region differences, at least for first filter (middle row of \fref{fig:gcnn_filter_viz}).
\fref{fig:rgcnn_filter_viz} shows the output of rGCNN stack for two trained models.
The featurization output (second row) appears to emphasize edge values more than the GCNN which make sense since once the graph is reduced there is no mechanism to hold these differences on the node feature vector aggregated to regions.
Also it is apparent that filters in the same level in the stack (same row in \fref{fig:gcnn_filter_viz}) are producing redundant output.
Finally, the two trained models, shown in \fref{fig:rgcnn_filter_viz}, with slightly different accuracies illustrate the redundancy the representation.
There are similarities between the outputs of the two models but it is hard to generate a simple, plausible explanation of how they are creating comparable representations.
This redundancy hinders interpretability.
Note that in both \fref{fig:gcnn_filter_viz} and \fref{fig:rgcnn_filter_viz} the filters are displayed in arbitrary order.

\section{Conclusion} \label{sec:conclusion}

In this work we presented a generalized multi-level GCNN architecture based predetermined clustering to reach a reduced graph and have shown that it can be the basis for effective models of microstructural properties and processes.
Given our results, the architecture should prove to be more efficient for large graphs and deep convolutional stacks than GCNNs operating solely on graphs of the data discretization.
The generality of the framework allows for more levels, more complex message passing between layers, and multiple levels of inputs than were studied in this work.

In preliminary studies deep, soft pooling such as \mincutpool\ and \diffpool\ appeared to not train well for the homogenization problem as we have posed it.
This may be due the fact that the adjacencies for our problems are determined by meshes whereas the clustering is determined by the initial data essentially independently of the mesh connectivity.
To enable soft pooling to work may be simply a matter of tailoring the auxiliary pooling to the homogenization task or at least ensuring the regression  objectives is primary.
A more hierarchical treatment may be necessary to efficiently process very large unstructured data and this will require learned pooling due the effects of convolution.
This endeavor is left for future work.
Pooling allows for construction of an encoder-decoder networks; however, presently effective decoders are limited~\cite{hamilton2017representation,gao2019graph}.

We also speculate that we could improved performance if we could we featurize edges information, \eg misorientation of neighboring grains, in a deep manner.
A filter acting on edge information, say $Z_A$ and $Z_B$ at neighboring nodes, would need to be permutationally invariant
\begin{equation}
w_1 Z_A + w_2 Z_B = w_1 Z_B + w_2 Z_A \ \Rightarrow
(w_1-w_2) Z_A + (w_2-w_1) Z_B = 0
\end{equation}
and hence $w_1 = w_2$ since $Z_A$ and $Z_B$ are arbitrary.
So filters based on edge data such as $w (Z_A + Z_B)$ or $w \, \abs(Z_A - Z_B)$ may boost accuracy.
Dual graphs based on the discretization vertices instead of the cells, together with alternative convolutions, may prove to be a richer embedding or at least one that can augment the cell-based graphs used in this work.

\section*{Acknowledgments}
This material is based upon work supported by the U.S. Department of Energy, Office of Science, Advanced Scientific Computing Research program.
Sandia National Laboratories is a multimission laboratory managed and operated by National Technology and Engineering Solutions of Sandia, LLC., a wholly owned subsidiary of Honeywell International, Inc., for the U.S. Department of Energy's National Nuclear Security Administration under contract DE-NA-0003525. This paper describes objective technical results and analysis. Any subjective views or opinions that might be expressed in the paper do not necessarily represent the views of the U.S. Department of Energy or the United States Government.



\appendix
\section{Appendix: Data generating models} \label{app:data_models}
\setcounter{equation}{0}
\renewcommand{\theequation}{\thesection.\arabic{equation}}

Details for the models and simulations that generated the three datasets described \sref{sec:data_sets} are given here.

\subsection{Polycrystal heat conduction}  \label{app:HF}

Steady heat conduction is governed by the partial differential equation
\begin{equation}
\grad \cdot \qb = \mathbf{0} \ ,
\end{equation}
where $\qb$ is the heat flux, expressed using the Fourier model
\begin{equation}
\qb = -\kappab \grad \theta.
\end{equation}
In the descrition below the temperature field $\theta$ is normalized.
In a crystal the conductivity $\kappab = \kappab(\phib(\Xb))$ can exhibit anisotropy \cite{graebner1992large,gofryk2014anisotropic,guo2015anisotropic} and in a polycrystal each crystal has a different orientation $\phib$.
The Kronecker product of the rotation $\Rb(\phib)$ with the conductivity tensor in the canonical orientation $\kappab = \sum_i \kappa_i \eb_i \otimes \eb_i$, gives the conductivity for a particular orientation:
\begin{equation}
\kappab(\phi) = \Rb(\phi) \boxtimes \kappab
= \sum_\alpha \kappa_\alpha \Rb(\phi) \eb_\alpha \otimes \Rb(\phi) \eb_\alpha \ .
\end{equation}

For our 2D demonstration we chose $\kappa_1, \kappa_2 =$ 1.0, 0.25 W/m-K.
The ratio of $\kappa_1$ and $\kappa_2$ is representative of non-metallic crystals.
For computational efficiency, we employed a simplified microstructure generation technique to create realizations of polycrystals in a square domain from Voronoi triangulation of random Latin Hypercube placements of vertices.
The triangular (crystal) regions were assigned crystal orientation $\phi_K$ from a uniform random distribution $\mathcal{U}(0,\pi)$ and then meshed with triangles.

Steady state solutions were obtained with standard linear triangular finite elements.
Dirichlet boundary conditions were applied to the left boundary, $\theta =0$, and right boundary, $\theta =1$, to establish a nominally uniform gradient.
The upper and lower boundaries were given homogeneous Neumann boundary conditions.
We computed the mean flux $\bar{\qb}$ to obtain an effective conductivity for the sample via:
\begin{equation}
\bar{\kappa} = \frac{L}{V \, \Delta \theta} \int_\Omega \qb(\Xb) \, \dV \ ,
\end{equation}
which is the output of interest.
Here $L=1$ is the distance between the left and right boundaries, $V=1$ is the area of the 2D sample, and $\Delta \theta = 1$ is the applied temperature difference.
Input features are the orientation angle $\phi_i$ and cell volume $\vol_i$ for all elements $i$.

\subsection{Crystal plasticity} \label{app:CP}

Polycrystalline aggregates where crystal plasticity governs the individual grain response is another common target for homogenization.
Here, the homogenized response of a representative sample of the grain structure characterized by $\phib(\Xb)$ is of interest.
Crystal plasticity, like most plasticity models, is comprised of an algebraic relation mapping recoverable elastic strain to stress and a set of ordinary differential equations governing the evolution of irrecoverable plastic strain.
Together, elastic and plastic strain comprise the total strain.

As in the heat flux exemplar in \sref{app:HF}, each sample is a collection of crystals assigned to convex sub-regions  $\Omega_K$ of a cube $\Omega$, and the number of crystals comprising the sample vary sample to sample.
DREAM3D \cite{dream3d} was used to create realistic 2D and 3D polycrystals' realizations.
The orientation $\phib_K$ of each grain was characterized by 3 angles for the 3D samples, while 1 angle was sufficient for the 2D samples.
These angle were drawn from a uniform (untextured) distribution of the appropriate special orthogonal group.
The 3D samples were created on a 25$\times$25$\times$25 grid, while the 2D samples were created on a 32$\times$32$\times$32 structured grid and then 2D slices were extracted with a spacing that ensured the 2D structures were uncorrelated.

The response of each crystal follows an elastic-viscoplastic constitutive relation \cite{taylor1934mechanism,kroner1961plastic,bishop1951xlvi,bishop1951cxxviii,mandel1965generalisation,dawson2000computational,roters2010overview}.
For the crystal elasticity, the (second Piola-Kirchhoff) stress $\Sb$ is given by
\begin{equation}
\Sb = \Sb(\Eb_e) =  \Cbb : \Eb_e \ ,
\end{equation}
where $\Cbb$ is the 4th order elastic modulus tensor and  $\Eb_e$ is elastic Lagrange strain.
In each sub-region $\Omega_K$, the orientation vector $\phib$ rotates $\Cbb = \Rb(\phib) \boxtimes \Cbb_0$ from the canonical $\Cbb_0$.
The face centered cubic (FCC) symmetry for each crystal was assumed, so that independent components of the elastic modulus tensor $\Cbb$ were: $C_{11}, C_{12}, C_{44} =$ 204.6, 137.7, 126.2 GPa which are representative of steel.
In each crystal, plastic flow can occur on any of the (rotated) 12 FCC slip planes where the plastic velocity gradient
\begin{equation}
\Lb_p = \Lb_p(\Sb) = \sum_\alpha \dot{\gamma}_{\alpha} (\Sb) \, \sb_\alpha \otimes \nb_\alpha
\end{equation}
is determined by slip rates  $\dot{\gamma}_{\alpha}$, slip directions $\sb_\alpha$, and slip plane normals $\nb_\alpha$.
The slip rate follows a power-law
\begin{equation}
\dot{\gamma}_{\alpha}=\dot{\gamma}_0\left|\frac{\tau_{\alpha}}{g_{\alpha}}\right|^{m-1}\tau_{\alpha} \ ,
\end{equation}
driven by the shear stress $\tau_\alpha = \sb_\alpha \cdot \Sb \nb_\alpha$ resolved on slip system $\alpha$, with parameters: reference slip rate $\dot{\gamma}_0$ = 1.0 s$^{-1}$, rate sensitivity exponent $m = 20$, and the initial slip resistance $g_{\alpha}$ = 122.0 MPa \cite{jones2018machine}.
The slip resistance evolves according to \cite{Kocks1976, mecking1976hardening} :
\begin{equation}
\dot{g}_\alpha = (H-R_d g_\alpha) \sum_\alpha |\dot{\gamma}_\alpha| \ ,
\end{equation}
with the hardening modulus $H = 355.0$ MPa and recovery constant $R_d = 2.9$.

Each sample was subject to tension at a strain rate of $1/s$ effected by minimal boundary conditions.
The overall response of the aggregate sample reflects the  anisotropy of each crystal which share boundaries and hence have kinematic constraints on their deformation.

\subsection{Viscoelastic matrix with stiff inclusions}  \label{app:VE}

To create realizations of a composite material, spheres of sizes drawn from a beta distribution and a uniform distribution of locations were packed into a cube.
The resulting distribution of volume fractions of inclusions were representative of a SVE.

The stiff inclusion were assigned the elastic properties of glass (Young's modulus 60 GPa and Poisson's ratio 0.33), while the matrix was taken to be silicone, a common engineering polymeric material.
We employed a Universal Polymer Model (UPM) \cite{adolf2009simplified,long2017linear} to model the silicone.
The UPM is a viscoelastic model for the Cauchy stress $\Tb$ of the hereditary integral type:
\begin{align}
\Tb &=
(K - K_\infty) \int_0^t \left( f_K(t-s) \tr \dot{\epsilonb} (s) \right) \, \mathrm{d}s \, \Ib + K_\infty \left( \tr \epsilonb \right) \, \Ib \\
&+ 2 (G - G_\infty) \int_0^t    \left( f_G(t-s) \dev \dot{\epsilonb} (s) \right) \, \mathrm{d}s + 2 G_\infty \dev \epsilonb  \nonumber
\end{align}
based on a bulk($K$)/shear($G$) split.
The strain measure $\epsilonb$ is given by the integration of the unrotated rate of deformation $\Db = \frac{1}{2} ( \grad_\xb \vb  + \grad^T_\xb \vb )$:
\begin{equation}
\epsilonb = \int_0^t \Rb^T(s) \Db(s) \Rb(s) \, \mathrm{d}s \ ,
\end{equation}
where $\Rb$ is the rotation tensor from the polar decomposition of the deformation gradient $\Fb$ and $\vb$ is velocity.
The relaxation kernels $f_K$ and $f_G$ are represented with Prony series with 20 relaxation times ranging from 1 $\mu$s to 3160 s.
The instantaneous bulk and shear moduli, $K=$ 920 MPa and $G=$ 0.362 MPa, and equilibrium bulk and shear moduli, $K_\infty=$ 920 MPa and $G_\infty$=0.084 MPa,  and all other parameters are given in \cref{long2017linear}.

As with the CP realizations, each composite sample was subject to tension at a strain rate of $6000/s$ effected by minimal boundary conditions.

\end{document}